\documentclass[pdflatex,sn-vancouver-num]{sn-jnl}

\usepackage{fix-cm}

\usepackage{hyperref}
\usepackage{float}
\usepackage{verbatim} 
\usepackage{apalike}
\usepackage{color}
\usepackage{tabularx}
\usepackage{threeparttable}
\usepackage{makecell}
\usepackage{longtable}
\usepackage{array}

\usepackage{amsmath}
\usepackage{bbm}
\usepackage{graphicx}
\usepackage{lineno}
\usepackage{graphicx}
\usepackage{dblfloatfix} 
\usepackage[dvipsnames]{xcolor}
\usepackage{soul}
\usepackage{color}
\usepackage{url}
\usepackage{float}
\usepackage{boldline}
\usepackage{xcolor,colortbl}
\usepackage{tikz}

\usepackage{xspace}

\usepackage{algorithm}
\usepackage{algpseudocode}

\usepackage{graphicx}
\usepackage{amsmath,amssymb,amsfonts}
\usepackage{amsthm}
\usepackage{mathrsfs}
\usepackage[title]{appendix}
\usepackage[dvipsnames,table]{xcolor}
\usepackage{textcomp}
\usepackage{manyfoot}
\usepackage{algorithm}
\usepackage{algorithmicx}
\usepackage{algpseudocode}
\usepackage{listings}

\usepackage{tikz}
\usepackage{pgfplots}
\usepackage{lineno}
\usepackage{dblfloatfix}
\usepackage{soul}
\usepackage{url}
\usepackage{float}
\usepackage{boldline}
\usepackage{pifont}
\usepackage{adjustbox}

\theoremstyle{thmstyleone}%
\theoremstyle{thmstyletwo}%

\theoremstyle{thmstylethree}%

\usepackage{bbm}
\usepackage{makecell}
\usepackage{tabularx}
\usepackage{booktabs}
\usepackage{array}
\usepackage{multirow}

\begin{document}

\title[Article Title]{Drift-Conditioned Structural Replay for Continual Cardiac Ultrasound Segmentation}


\author[1]{\fnm{Zahid} \sur{Ullah}}\email{zahid1989@dongguk.edu}

\author[1]{\fnm{Jaeyoon} \sur{Seo}}\email{tonyno193@gmail.com}

\author*[1]{\fnm{Jihie} \sur{Kim}}\email{jihie.kim@dgu.edu}

\affil*[1]{\orgdiv{Department of Computer Science and Artificial Intelligence}, \orgname{Dongguk University}, 
\city{Seoul}, \postcode{04620}, \country{South Korea}}




\abstract{
Conventional experience replay typically applies fixed preservation constraints to historical exemplars, overlooking variations in how learned representations, predictions, and anatomical structures change during sequential training. To address this limitation, we propose Drift-Conditioned Structural Replay Loss (DCSRL), an exemplar-adaptive supervision mechanism that regulates anatomical preservation according to source-referenced knowledge drift. DCSRL operates within a reliability-guided replay framework that selects informative historical exemplars using anatomical content and boundary characteristics while adaptively regulating replay during training. For each retained exemplar, DCSRL constructs a frozen source-task reference signature and subsequently measures changes in feature representations, class probabilities, boundary responses, and predicted anatomical structures. These complementary drift measurements are aggregated and normalized to generate bounded, exemplar-specific coefficients that dynamically adjust boundary and structural replay losses without modifying the underlying memory-selection policy. We evaluate DCSRL under sequential learning across CAMUS, CardiacNet, and EchoNet-Dynamic, comparing its performance with regularization-, distillation-, and replay-based continual-learning approaches. Experimental findings indicate that DCSRL improves overall final segmentation performance and anatomical boundary preservation, particularly under heterogeneous imaging conditions, although trade-offs remain in knowledge transfer across tasks. 
These findings suggest that source-referenced multidimensional drift conditioning offers a promising direction for adaptive anatomical knowledge preservation in continual cardiac ultrasound segmentation.
}




\maketitle

\section{Introduction}
\label{sec:introduction}

\begin{figure}[!ht]
    \centering
    \includegraphics[width=0.6\columnwidth]{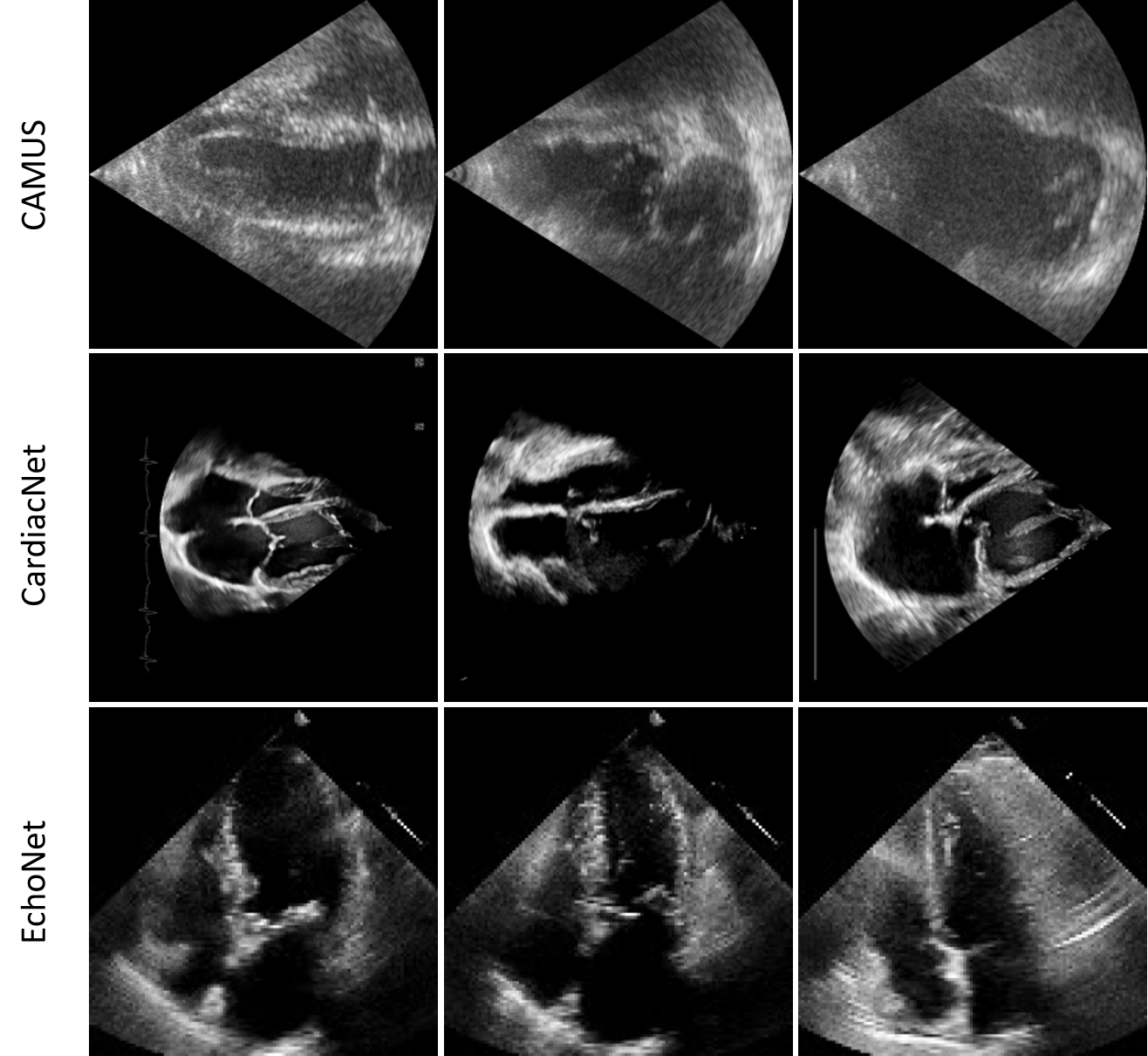}
    \caption{Representative echocardiographic images from CAMUS, CardiacNet, and EchoNet-Dynamic.}
    \label{Fig1}
\end{figure}

Echocardiographic measurements of cardiac structure and function depend
on accurate delineation of structures such as the left ventricular (LV)
cavity, myocardium, and left atrium. Maintaining segmentation accuracy
across datasets is challenging because both image appearance and the
available annotations can change
\cite{leclerc2019deep,deng2024memsam,ouyang2020video}. In the sequence
studied here, CAMUS provides annotations of the LV cavity, myocardium,
and left atrium in two- and four-chamber views, CardiacNet provides
chamber annotations, and EchoNet-Dynamic provides LV-only supervision.
During the final task, the model must therefore adapt to a new image
distribution while retaining its ability to delineate structures for
which it no longer receives direct supervision.
Figure~\ref{Fig1} illustrates the variation in image appearance across
the three datasets.

Continual learning seeks to preserve performance on earlier tasks
during sequential training
\cite{parisi2019continual,de2021continual,wu2024continual}.
Elastic Weight Consolidation penalizes changes to parameters important
for earlier tasks \cite{kirkpatrick2017overcoming}, while Learning
without Forgetting constrains changes in predictions through
distillation \cite{li2017learning}. Experience replay revisits a
bounded set of historical examples during training. In medical imaging,
replay methods have addressed how to populate this memory by retaining
diverse cases \cite{perkonigg2021dynamic},
atypical examples \cite{bera2023memory}, or examples with informative
anatomical boundaries \cite{ullah2026bbr}. Once an exemplar is sampled,
a further question arises: how much anatomical supervision does it
require at the current stage of training?

\begin{figure}[!t]
    \centering
    \includegraphics[width=\columnwidth]{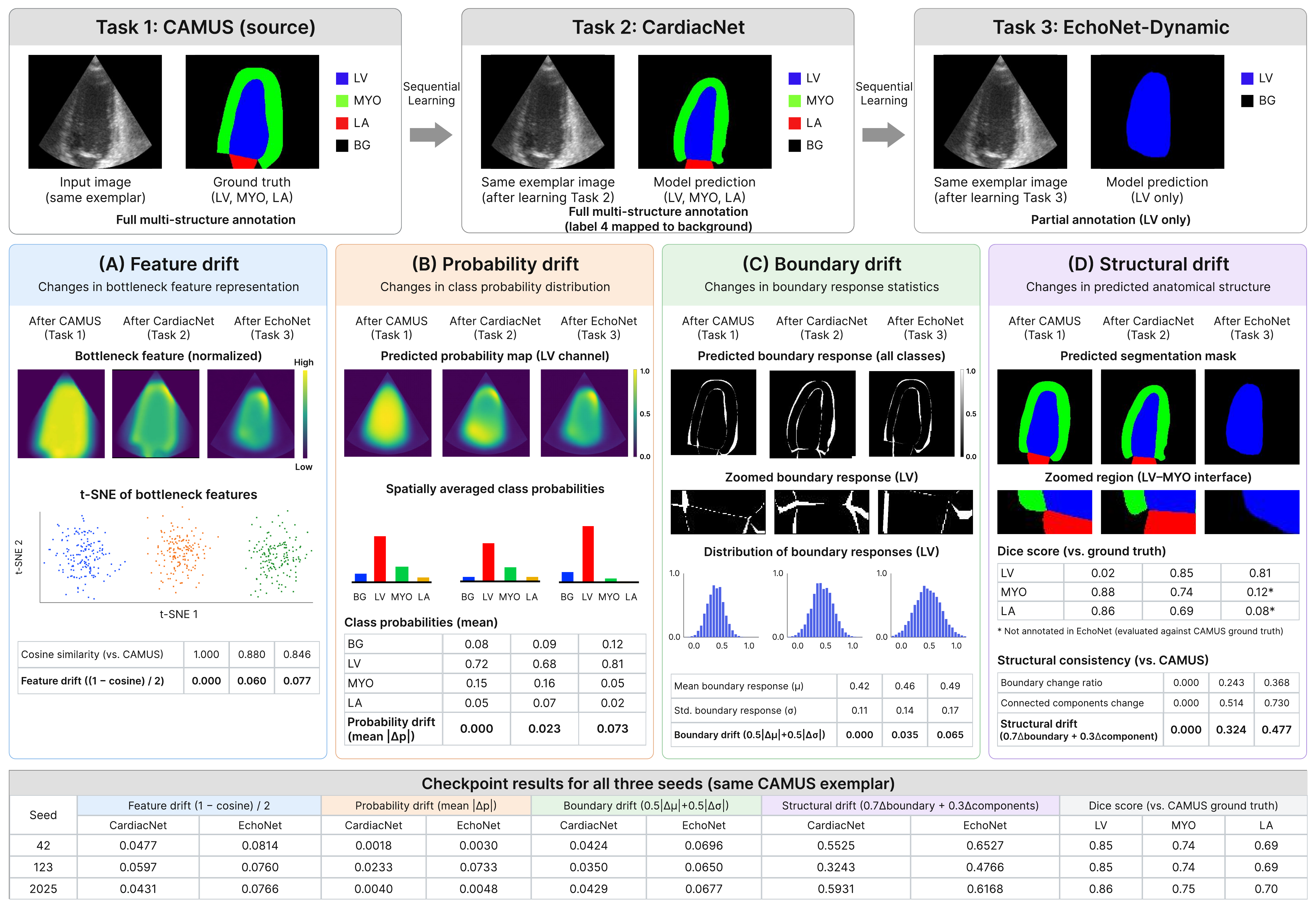}
    
\caption{Source-referenced knowledge drift across sequential
echocardiographic tasks.
The same CAMUS exemplar is evaluated after training
on CAMUS, CardiacNet, and EchoNet-Dynamic.
Panels (A)-(D) illustrate changes in bottleneck
features, class probabilities, boundary responses,
and predicted anatomical structures, respectively.
All drift measurements are referenced to the
model state after CAMUS training.}
    \label{Fig11}
\end{figure}

Consider a CAMUS exemplar whose predicted myocardial wall develops a
small break after training on CardiacNet. The break may affect few
pixels and leave regional overlap high, yet it changes the connectivity
of the predicted myocardium. Another exemplar may retain an accurate
contour despite a change in its latent representation. Such cases
motivate measuring changes at several levels and using boundary
information alongside region-based supervision
\cite{kervadec2021boundary}. We use \emph{drift} to denote a change in
the model's response to the same exemplar relative to its source-task
state. Figure~\ref{Fig11} illustrates how these responses evolve over
successive tasks.

We propose \textbf{Drift-Conditioned Structural Replay Loss (DCSRL)},
which adjusts anatomical supervision for each replayed exemplar. Its
underlying \emph{reliability-guided replay} baseline considers
anatomical class content and boundary complexity when retaining
exemplars and adjusts the overall weight of replay during training.
DCSRL uses the same selection, replacement, and sampling rules. For
each retained exemplar, it records a frozen source-task signature
comprising a normalized bottleneck representation, probability
statistics for the supervised classes, boundary-response statistics,
and a predicted anatomical mask. When the exemplar is sampled again,
DCSRL compares its current responses with this signature. It combines
and normalizes the measured changes into bounded weights that increase
boundary and structural supervision as drift grows. This requires
additional storage and computation during training and introduces no
additional trainable parameters or inference operations. 
The comparison with reliability-guided replay isolates the contribution
of drift-conditioned supervision under a common memory policy.
The experiments address three questions.

\begin{itemize}
    \item \textbf{Does drift conditioning improve final segmentation?}
    DCSRL achieved the highest final average accuracy, defined as mean
    final Dice across the three datasets
    ($0.8784 \pm 0.0017$, compared with $0.8718 \pm 0.0029$ for
    reliability-guided replay). It improved the mean result in 14 of
    15 dataset--metric combinations spanning regional overlap,
    boundary agreement, and surface distance.

    \item \textbf{How do the final gains relate to transfer?}
    DCSRL achieved lower backward and forward transfer scores than
    reliability-guided replay. Reporting both measures distinguishes
    final segmentation performance from the effects of learning across
    tasks.

    \item \textbf{Which components contribute to anatomical
    preservation?} We ablated each of the four drift signals and each
    of the two conditioning pathways. The complete objective yielded
    the best mean result among the tested variants on all five
    EchoNet-Dynamic metrics, while several ablated variants achieved
    lower surface-distance errors on CAMUS.
\end{itemize}

\section{Related Work}
\label{sec:related_work}

This section reviews four lines of work that bear on DCSRL:
echocardiographic segmentation and the datasets that define the
present task sequence; continual learning and its relation to drift in
non-stationary data; replay-based continual learning, with emphasis on
how memory is populated, retrieved, and supervised; and loss functions
that preserve boundary, shape, and topology. The section closes by
positioning DCSRL relative to these methods.

\subsection{Deep Learning for Echocardiographic Segmentation}
\label{subsec:echo_segmentation}

Accurate cardiac segmentation requires both global context to identify
a chamber and local detail to delineate its contour. U-Net
\cite{ronneberger2015u} combines these sources of information through
skip connections between its encoder and decoder, and nnU-Net
\cite{isensee2021nnu} showed that a carefully configured U-Net remains a
strong baseline across many medical segmentation benchmarks. Attention
U-Net \cite{oktay2018attention} uses attention gates to emphasize
spatially relevant features, while TransUNet \cite{chen2021transunet}
and Swin-Unet \cite{cao2022swin} incorporate transformer-based
contextual modeling. Large pretrained segmenters such as the Segment
Anything Model \cite{kirillov2023segment} and its medical adaptation
MedSAM \cite{ma2024segment} have broadened the available backbones, and
MemSAM \cite{deng2024memsam} adapted this family to echocardiography
videos through a temporal memory mechanism.

In echocardiography, CAMUS \cite{leclerc2019deep} established a
benchmark for segmenting the left ventricular (LV) cavity, myocardium,
and left atrium in two- and four-chamber views, and LU-Net
\cite{leclerc2020lu} improved robustness on this benchmark with a
multistage attention design. EchoNet-Dynamic \cite{ouyang2020video}
demonstrated video-based LV segmentation and estimation of cardiac
function at a larger scale, and CardiacNet \cite{yang2024cardiacnet}
introduced echocardiographic data for cardiac disease assessment.
Several studies have also shown that overlap-optimized networks can
produce anatomically implausible masks, such as a myocardium with a
gap or a cavity split into several regions. Anatomically constrained
networks \cite{oktay2017anatomically} regularize predictions with a
learned shape representation, and Painchaud et al. enforced anatomical
plausibility \cite{painchaud2020cardiac} and temporal consistency
\cite{painchaud2022echocardiography} in cardiac segmentation. These
studies treat anatomical plausibility as a property that overlap
metrics can miss, which motivates the boundary and surface measures
used in this work.

These datasets differ in annotation scope, which matters when they are
encountered sequentially. An LV tracing provides a target for updating
the ventricular contour, but the effect of that update on a previously
learned myocardial boundary can be assessed only on an earlier task.
Existing echocardiographic methods are trained and evaluated on a fixed
dataset; they do not address how a model's anatomy changes when
datasets with different label spaces arrive in sequence.

\subsection{Continual Learning and Drift in Non-Stationary Data}
\label{subsec:continual_medical_segmentation}

Continual learning addresses training on a sequence of data
distributions while limiting catastrophic forgetting of earlier ones
\cite{parisi2019continual,de2021continual,wang2024comprehensive}.
Clinical imaging data have this character: scans accumulate over time
and across institutions, scanners, and annotation protocols, so a
deployed model is repeatedly exposed to new data rather than to a
single fixed corpus \cite{alzubaidi2021review}. The data-stream
literature describes such changes as concept drift, a change in the
joint distribution of inputs and targets over time, and proposes
methods to detect and adapt to it \cite{gama2014survey,lu2018learning}.
DCSRL uses \emph{drift} in a narrower, model-centered sense: a change
in the model's own response to a fixed stored exemplar relative to the
model state at which that exemplar was learned. The quantity is
measured per exemplar rather than per data stream, and it is used to
set supervision strength rather than to trigger model adaptation.

Continual-learning methods differ in how they preserve earlier
knowledge during model updates. Regularization methods estimate the
importance of each parameter and penalize changes to important ones:
Elastic Weight Consolidation (EWC) \cite{kirkpatrick2017overcoming}
uses Fisher information, Synaptic Intelligence
\cite{zenke2017continual} accumulates each parameter's contribution to
the loss along the training trajectory, and Memory Aware Synapses
\cite{aljundi2018memory} uses the sensitivity of the learned function.
Learning without Forgetting (LwF) \cite{li2017learning} instead uses
predictions from an earlier model as distillation targets on incoming
data. For dense prediction, the choice of what to preserve also has a
spatial dimension. Continual semantic segmentation methods distill
intermediate features \cite{michieli2019incremental}, regularize
latent representations \cite{michieli2021continual}, model the
shifting meaning of the background class \cite{cermelli2020modeling},
or apply multi-scale pooled distillation with pseudo-labeling
\cite{douillard2021plop}. A parameter-importance estimate does not
locate a displaced contour, and agreement on incoming images provides
only indirect evidence of performance on earlier anatomy.

In medical image segmentation, early studies showed that sequential
fine-tuning across sites or anatomies causes substantial forgetting and
evaluated adaptations of EWC and distillation
\cite{baweja2018towards,ozdemir2018learn,gonzalez2025wrong}. Lifelong
nnU-Net \cite{gonzalez2023lifelong} standardized the training and
evaluation of continual segmentation, including a cardiac use case,
and surveys have organized the growing literature on continual medical
image analysis \cite{wu2024continual}. Class-incremental methods for
CT have used pseudo-labels and class-specific output heads to add
organs and tumors without access to earlier data
\cite{zhang2023continual,ji2023continual}. Continual learning has also
been applied to cardiac signals across diseases, time, modalities, and
institutions \cite{kiyasseh2021clinical} and to domain shifts in chest
X-ray classification \cite{lenga2020continual}. These studies
establish that clinical models must accommodate both new domains and
new label spaces, the combination encountered in the present
CAMUS $\rightarrow$ CardiacNet $\rightarrow$ EchoNet-Dynamic sequence.

Several methods incorporate shape or morphology explicitly. Shape and
Semantics-based Selective Regularization (S$^{3}$R) uses shape and
semantic information to identify parameters to protect during
subsequent training \cite{zhang2023s}. CLMS \cite{li2025clms}
addresses source-free adaptation through multi-scale reconstruction
and style alignment to preserve morphology as image appearance
changes. Work on online continual learning \cite{caccia2021new} has
also documented abrupt changes in learned representations when new
classes are introduced. For segmentation, however, feature displacement
and anatomical error are distinct quantities: a representation can
change while its decoded contour remains stable, and a contour can
break while overall overlap remains high. This observation motivates
measuring drift at several levels rather than relying on a single
representation-level signal.

\subsection{Replay-Based Continual Learning}
\label{subsec:replay_continual_learning}

Experience replay \cite{rolnick2019experience,chaudhry2019tiny}
interleaves historical examples with incoming data, and even small
episodic memories can support retention. Replay has also been used as
a constraint rather than a training signal: Gradient Episodic Memory
\cite{lopez2017gradient} and A-GEM \cite{chaudhry2018efficient}
project updates so that the loss on stored examples does not increase.
Generative replay \cite{shin2017continual} replaces stored images with
samples from a generative model, at the cost of training and
maintaining that model.

With a limited memory, both buffer composition and retrieval determine
which observations contribute to training. iCaRL
\cite{rebuffi2017icarl} selects exemplars that approximate class means
in feature space and combines rehearsal with distillation. Gradient-based
sample selection \cite{aljundi2019gradient} favors stored examples
with diverse gradients, Rainbow Memory \cite{bang2021rainbow} favors
diverse and uncertain samples, and GDumb \cite{prabhu2020gdumb} showed
that a class-balanced memory alone forms a strong baseline. For dense
prediction, Zhu et al. \cite{zhu2023continual} learned an automatic
memory-sample selection policy. Medical imaging methods have used
acquisition-style diversity \cite{perkonigg2021dynamic}, sample
atypicality \cite{bera2023memory}, and the combination of boundary
information and class balance in BBR-Net \cite{ullah2026bbr} to guide
retention. Atlas Replay \cite{ranem2024continual} takes a different
approach to memory representation, using anatomical prototypes and
registration for continual prostate MRI segmentation.

A separate group of methods changes what a retrieved exemplar
contributes. Maximally Interfered Retrieval (MIR)
\cite{aljundi2019online} estimates which stored examples would incur
the largest loss increase after a prospective update and gives those
examples retrieval priority. Dark Experience Replay (DER)
\cite{buzzega2020dark} stores historical logits and penalizes
discrepancies between those logits and current outputs, and DER++
adds supervised rehearsal on stored labels. X-DER
\cite{boschini2022class} revises stored logits as training proceeds,
and CLS-ER \cite{arani2022learning} distills from slowly and rapidly
updated model averages. MIR uses a prospective change to guide
retrieval, whereas DER and its extensions use a historical or
consolidated prediction as a training target. These precedents
distinguish the decision to retrieve an exemplar from the choice of
supervision it receives once retrieved. DCSRL addresses the second
decision: it leaves retrieval unchanged and uses a measured change in
the exemplar's response to set the strength of anatomical
supervision, while the stored annotation, not a stored prediction,
supplies the target.

\subsection{Anatomical and Structural Knowledge Preservation}
\label{subsec:structural_preservation}

Two masks with similar Dice scores can differ in contour placement or
in the continuity of a thin myocardial wall. Region-based and
distribution-based losses weight these errors differently
\cite{ma2021loss}. Boundary loss
\cite{kervadec2021boundary} incorporates distance
to the reference contour into the training objective.
Hausdorff-inspired losses give greater weight to large spatial
discrepancies \cite{karimi2019reducing}, while boundary-wise objectives
provide additional supervision at anatomical interfaces
\cite{lin2024boundary}. Kang et al. \cite{kang2024advancing}
incorporated neighborhood-based label smoothing and reverse boundary
attention into Diffusion Transformer Segmentation to improve
morphological representation and boundary delineation.

Topology-aware objectives address connectivity directly. Hu et al.
\cite{hu2019topology} matched the persistent homology of predicted and
reference likelihood maps, Clough et al. \cite{clough2020topological}
used persistent homology to impose prior topological constraints, and
clDice \cite{shit2021cldice} preserves the connectivity of tubular
structures through a skeleton-based overlap measure. These methods
establish ways to penalize spatial and topological errors in a static
training setting. In a continual setting, a further design choice is
how strongly to apply such penalties to a particular replayed image.
A uniform weight applies the same supervision to an exemplar whose
myocardial wall has broken and to one whose prediction is unchanged.

\subsection{Positioning of DCSRL}
\label{subsec:positioning}

DCSRL addresses this choice within a reliability-guided replay
baseline that considers anatomical class content and boundary
complexity when retaining exemplars and adjusts the overall replay
weight during training. After an exemplar is sampled, DCSRL compares
its current features, supervised-class probabilities, boundary
responses, and predicted mask with a frozen signature recorded during
its source task. The measured changes determine bounded,
exemplar-specific weights for the boundary and structural replay
losses. Using the same memory selection, replacement, and sampling
rules in both methods allows the comparison to isolate the effect of
conditioning anatomical supervision on the exemplar's observed drift.

\begin{table*}[!t]
\centering
\caption{Comparison of representative continual-learning
approaches relevant to medical image segmentation.}
\label{tab:related_work_comparison}

\footnotesize
\setlength{\tabcolsep}{3pt}
\renewcommand{\arraystretch}{1.02}

\begin{tabularx}{\textwidth}{
    >{\raggedright\arraybackslash}p{2.1cm}
    >{\raggedright\arraybackslash}p{2.5cm}
    >{\raggedright\arraybackslash}X
}
\toprule
\textbf{Category} &
\textbf{Method} &
\textbf{Relation to DCSRL} \\
\midrule

Regularization &
EWC \cite{kirkpatrick2017overcoming} &
Protects important parameters; DCSRL adapts anatomical replay losses
using measured drift. \\

Distillation &
LwF \cite{li2017learning} &
Preserves previous predictions on incoming data; DCSRL measures
feature, probability, boundary, and structural drift on stored
exemplars. \\

Segmentation distillation &
PLOP \cite{douillard2021plop} &
Distills multi-scale features on incoming images; DCSRL conditions
supervision on replayed exemplars. \\

Benchmark framework &
Lifelong nnU-Net \cite{gonzalez2023lifelong} &
Standardizes continual segmentation with uniform replay and
regularization; DCSRL adds exemplar-specific supervision weights. \\

Exemplar replay &
iCaRL \cite{rebuffi2017icarl} &
Selects representative exemplars; DCSRL adjusts supervision after
memory retrieval. \\

Interference replay &
MIR \cite{aljundi2019online} &
Prioritizes samples with predicted interference; DCSRL conditions
replay losses on observed drift. \\

Logit replay &
DER/DER++ \cite{buzzega2020dark} &
Preserves stored logits; DCSRL keeps the annotation as the target and
adapts boundary and anatomical supervision. \\

Dynamic memory &
Perkonigg et al. \cite{perkonigg2021dynamic} &
Maintains diverse historical samples; DCSRL measures changes in
retained knowledge. \\

Anatomy-aware replay &
BBR-Net \cite{ullah2026bbr} &
Selects samples using class and boundary information; DCSRL adapts
structural preservation strength. \\

Reliability-guided replay &
Reliability-Guided Replay &
Uses reliability-based selection and adaptive replay; DCSRL adds
exemplar-specific drift conditioning. \\

\midrule

\textbf{Drift-conditioned replay} &
\textbf{DCSRL (Ours)} &
\textbf{Combines four source-referenced drift measures to adapt
boundary and structural replay losses without changing memory
selection.} \\

\bottomrule
\end{tabularx}
\end{table*}

Table~\ref{tab:related_work_comparison} summarizes representative
continual-learning strategies relevant to medical image segmentation
and their differences from DCSRL. Existing approaches primarily
preserve important parameters, distill historical predictions or
features, or improve the construction and retrieval of replay memory.
Static boundary and topology losses, in turn, apply a fixed weight to
every training image. Although these strategies mitigate forgetting or
improve contour quality, they do not regulate anatomical preservation
according to multiple source-referenced drift measurements for each
replayed exemplar. DCSRL addresses this complementary problem by
measuring feature, probability, boundary, and structural changes and
using their normalized discrepancies to adapt the strength of boundary
and anatomical supervision. Because this conditioning operates after
exemplar selection, it can be combined with the memory-selection and
retrieval strategies reviewed above rather than replacing them.

\section{Proposed Method}
\label{sec:method}

\begin{figure*}[!t]
    \centering
    \includegraphics[width=\textwidth]{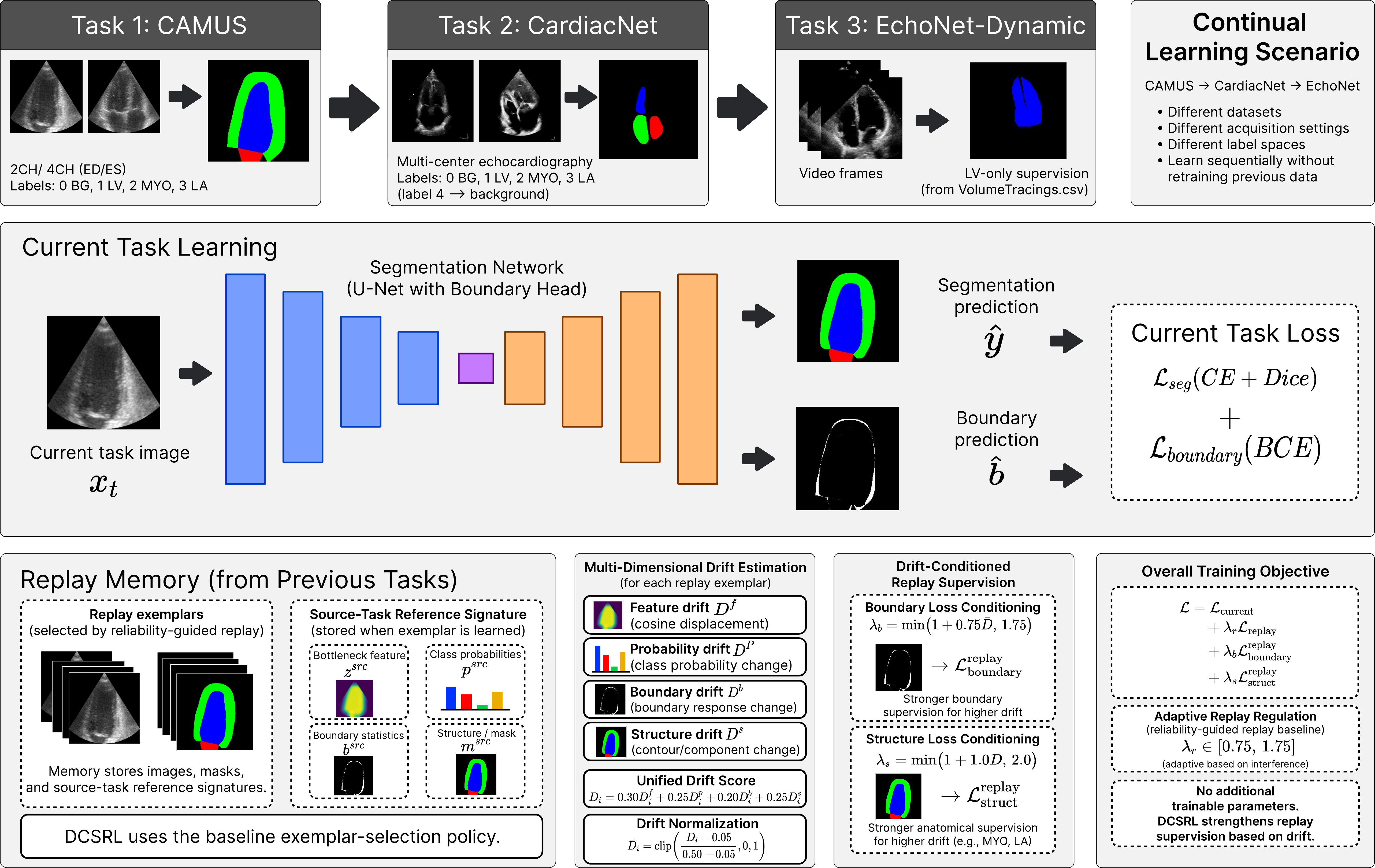}
    \caption{DCSRL training procedure.
Current-task images receive segmentation and boundary
supervision, while historical exemplars are retrieved
using a reliability-guided replay policy. For each
replayed exemplar, feature, probability, boundary,
and structural drift are measured relative to its
frozen source-task reference signature. These
measurements determine exemplar-specific boundary
and structural loss weights without modifying the
underlying memory-selection policy.}
    \label{fig2}
\end{figure*}

Drift-Conditioned Structural Replay Loss (DCSRL) adjusts anatomical
supervision for each replayed exemplar according to how the model's
response has changed since the exemplar was learned. Each retained
image has a frozen source-task signature. During replay, DCSRL compares
the current response with this signature and uses the measured drift
to weight boundary and structural losses. Figure~\ref{fig2} shows
where these weights enter the training objective.

\subsection{Sequential Segmentation and Network Outputs}
\label{subsec:problem}

We train a single segmentation network on three successive tasks,
\begin{equation}
    \mathrm{CAMUS}
    \rightarrow \mathrm{CardiacNet}
    \rightarrow \mathrm{EchoNet\text{-}Dynamic}.
    \label{eq:task_sequence}
\end{equation}
Task $t$ provides a dataset
$\mathcal{D}_t=\{(\mathbf{x}_n^t,\mathbf{y}_n^t,m_n^t)\}_{n=1}^{N_t}$,
where $\mathbf{x}_n^t$ is an image, $\mathbf{y}_n^t$ is its target
mask, and $m_n^t$ identifies the annotation mode used for training.
Earlier training images are accessible through a bounded replay memory
$\mathcal{M}_{t-1}$. We denote the foreground channels used for
exemplar $i$ by $\mathcal{C}_i$.

The shared output space comprises background (BG), left ventricular
cavity (LV), myocardium (MYO), and left atrium (LA). CAMUS provides
annotations for all three foreground structures \cite{leclerc2019deep}.
CardiacNet provides chamber annotations in its native label space.
EchoNet-Dynamic provides LV tracings \cite{ouyang2020video}. For
LV-only supervision, the target distinguishes LV from its complement
without assigning a specific anatomical class to non-LV pixels.

\begin{figure*}[!t]
    \centering
    \includegraphics[width=\textwidth]{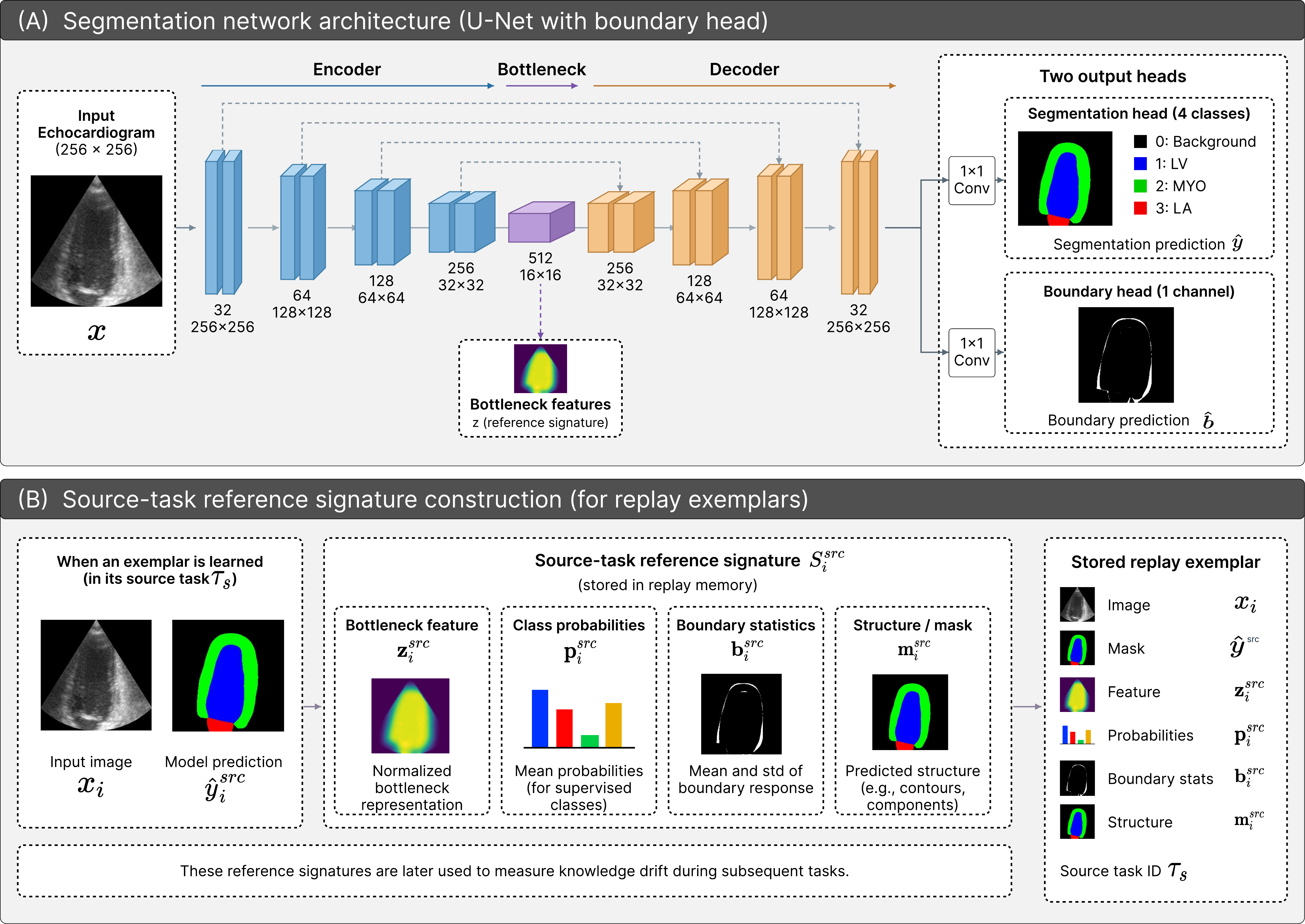}
    \caption{Network architecture and source-task reference construction.
(A) The U-Net backbone produces segmentation predictions,
auxiliary boundary responses, and bottleneck features.
(B) For each retained exemplar, a source-task reference
signature is constructed from its normalized bottleneck
representation, spatially averaged class probabilities,
boundary-response statistics, and predicted segmentation
mask. The stored reference remains frozen during
subsequent continual-learning tasks.}
    \label{fig3}
\end{figure*}

The backbone is a U-Net \cite{ronneberger2015u} with four encoder
stages, a bottleneck, and four decoder stages connected to the encoder
through skip connections (Fig.~\ref{fig3}A). For image $\mathbf{x}_i$,
the network returns
\begin{equation}
    f_\theta(\mathbf{x}_i)
    =(\mathbf{S}_i,\mathbf{B}_i,\mathbf{z}_i),
    \label{eq:model_output}
\end{equation}
where $\mathbf{S}_i\in\mathbb{R}^{4\times H\times W}$ contains the
segmentation logits, $\mathbf{B}_i\in\mathbb{R}^{1\times H\times W}$
contains the auxiliary boundary logits, and $\mathbf{z}_i$ is the
globally pooled bottleneck representation. We write
$\mathbf{p}_i=\operatorname{softmax}(\mathbf{S}_i)$ for the class
probabilities and $\mathbf{Q}_i=\operatorname{sigmoid}(\mathbf{B}_i)$
for the boundary response.

\subsection{Reliability-Guided Replay and Source References}
\label{subsec:memory}

\subsubsection{Reliability-guided replay baseline}
\label{subsubsec:reliability_guided_replay}

We use reliability-guided replay as the reference configuration for
DCSRL. The baseline maintains a bounded memory of historical images
and their segmentation masks. A priority score governs both exemplar
retention and replay sampling. During training, an epoch-level
controller adjusts the overall replay weight, and a structural loss
provides additional supervision on eligible replay masks
(Section~\ref{subsec:adaptive_replay}).

\subsubsection{Reliability-score definitions}\label{app:replay_scores}

Let $h(z)=\min(1,\max(0,z))$. For an image $x$ normalized to
$[0,1]$, the image reliability score is
\begin{equation}
\begin{aligned}
r^{\mathrm{img}} ={}&
  0.4h\!\left(\frac{\operatorname{std}(x)}{0.25}\right)
 +0.4h\!\left(\frac{\operatorname{var}(\Delta x_8)}{500}\right)
 &+0.2h\!\left(\frac{\operatorname{mean}[x>0.05]}{0.60}\right),
\end{aligned}
\end{equation}
where $x_8$ is $255x$ converted to an unsigned 8-bit image and
$\Delta$ is the OpenCV Laplacian with its default kernel.

Let $A$ be the number of foreground pixels and $P$ the number of
boundary pixels. A pixel is marked as a boundary pixel if any
horizontal or vertical neighbor has a different label. This rule
includes interfaces between foreground classes and marks both sides
of a transition. For an $H\times W$ mask, $b=P/(HW)$. We set
$r^{\mathrm{anat}}=0$ when $A<10$. Otherwise,
\begin{equation}
r^{\mathrm{anat}}
=0.4a
+0.4h\!\left(\frac{4\pi A}{P^2+10^{-8}}\right)
+0.2h\!\left(\frac{1}{1+P/(\sqrt{A}+10^{-8})}\right),
\end{equation}
where $a=1$ if $0.005\leq A/(HW)\leq0.55$ and $a=0.4$ otherwise.

For EchoNet-Dynamic, temporal scores are computed from traced LV
areas in the original frame coordinates. A video with one traced
frame receives a score of 0.5. For a video with multiple traced
frames, a frame with area at most 10 pixels receives 0.1. For the
remaining frames, the score is 1.0 when the ratio of the minimum to
maximum traced area lies in $[0.20,0.95]$, and 0.5 otherwise.

\paragraph{Exemplar priority.}
For candidate image $x_i$ and its training mask $y_i$, we compute
\begin{equation}
    p_i = 0.15c_i + 0.20b_i + 0.20r_i^{\mathrm{img}}
        + 0.25r_i^{\mathrm{anat}} + 0.20r_i^{\mathrm{temp}}.
    \label{eq:reliability_replay_priority}
\end{equation}
Here, $c_i$ is the fraction of the three foreground label indices
present in $y_i$, and $b_i$ is the fraction of pixels adjacent to
a horizontal or vertical label transition. The image score
$r_i^{\mathrm{img}}$ combines contrast, sharpness, and visibility.
The anatomical score $r_i^{\mathrm{anat}}$ combines foreground area,
compactness, and boundary smoothness. The temporal score
$r_i^{\mathrm{temp}}$ is fixed at 0.5 for CAMUS and CardiacNet and
is derived from traced LV areas for EchoNet-Dynamic. The component
definitions are given in Appendix~\ref{app:replay_scores}.

\paragraph{Memory construction.}
After each task, we score candidates from its training loader, build
a task-specific buffer, and merge its retained entries into the
global buffer. Both buffers have a capacity of 1800 exemplars.
Candidates are appended while space is available. For a full buffer,
the target task fractions are proportional to 0.25, 0.50, and 0.25
for CAMUS, CardiacNet, and EchoNet-Dynamic, respectively, and are
normalized over the tasks represented at that stage.

If the incoming task is below its target, we consider the stored
task with the largest excess over its target for replacement.
Otherwise, we consider entries from the incoming task itself.
The incoming exemplar replaces the lowest-priority entry in the
selected task only if its priority is strictly higher. Thus, the
final targets of 450, 900, and 450 guide replacement, while the
realized counts depend on candidate availability and priority.

\paragraph{Replay sampling.}
Each replay batch contains eight draws with replacement from the
entire global buffer. For buffer $\mathcal{M}$, the probability
of drawing exemplar $i$ is
\begin{equation}
    \Pr(i\mid\mathcal{M})
    = \frac{\max(p_i,10^{-8})}
           {\sum_{j\in\mathcal{M}}\max(p_j,10^{-8})}.
    \label{eq:reliability_replay_sampling}
\end{equation}
Both methods use this distribution. DCSRL additionally stores a
source-task signature and uses it to adjust supervision after an
exemplar has been sampled.

\noindent\textbf{Source-task reference signatures.}
For each exemplar retained in memory, DCSRL records
a frozen source-task reference signature comprising
its normalized bottleneck representation,
class-probability statistics, boundary-response
statistics, and predicted segmentation mask:

\begin{equation}
    \mathcal{R}_i =
    \left(
        \hat{\mathbf{z}}_i^0,\,
        \boldsymbol{\mu}_i^{p,0},\,
        \mu_i^{b,0},\,
        \sigma_i^{b,0},\,
        \hat{\mathbf{Y}}_i^0
    \right).
    \label{eq:reference_signature}
\end{equation}
The superscript $0$ denotes the source state of that particular
exemplar. For either the source state ($a=0$) or the current state
($a=t$), the signature quantities are
\begin{align}
    \hat{\mathbf{z}}_i^a
    &= \frac{\mathbf{z}_i^a}{\|\mathbf{z}_i^a\|_2+\epsilon},
    \label{eq:normalized_feature}\\
    \mu_{i,c}^{p,a}
    &= \frac{1}{HW}\sum_{u,v}p_{i,cuv}^a,
    \qquad c\in\mathcal{C}_i,
    \label{eq:prob_reference}\\
    \mu_i^{b,a}
    &= \operatorname{mean}(\mathbf{Q}_i^a),
    \qquad
    \sigma_i^{b,a}=\operatorname{std}(\mathbf{Q}_i^a),\\
    \hat{Y}_{i,uv}^a
    &= \arg\max_c p_{i,cuv}^a .
\end{align}
Here, $\epsilon>0$ is a numerical stabilizer. The stored signature
remains fixed throughout subsequent tasks. Its mask records the
model's source prediction. The exemplar's annotation supplies the
target for the supervised replay losses.

\subsection{Exemplar-Level Drift}
\label{subsec:multidrift}

\begin{figure*}[!t]
    \centering
    \includegraphics[width=0.9\textwidth]{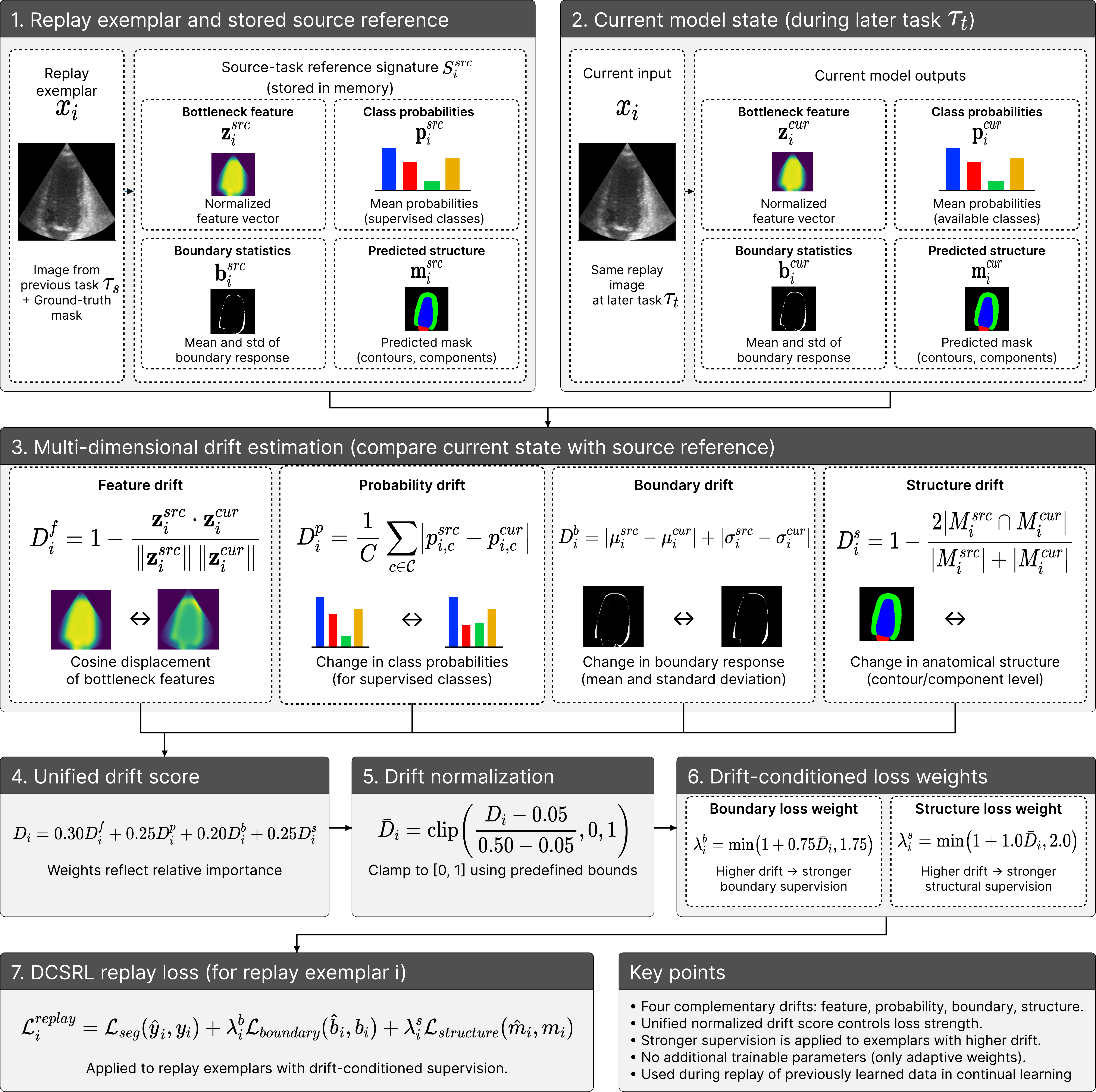}
    \caption{Multidimensional drift estimation and adaptive replay supervision.
For each replayed exemplar, feature, probability, boundary,
and structural discrepancies are measured relative to
its frozen source-task reference signature.
The four drift measurements are combined into an
aggregate score and normalized to derive
exemplar-specific boundary and structural loss
weights. These weights regulate anatomical
supervision during continual training.}
    \label{fig4}
\end{figure*}

For each sampled exemplar, we measure changes in its representation,
class probabilities, boundary response, and predicted structure
(Fig.~\ref{fig4}).

\paragraph{Feature and probability drift.}
Feature drift measures the angular displacement of the bottleneck
representation. Probability drift measures the mean absolute change
in spatially averaged probabilities over the exemplar's supervised
foreground channels:
\begin{align}
    D_i^f
    &= \operatorname{clip}\!\left(
        \frac{1-\langle\hat{\mathbf{z}}_i^t,
                       \hat{\mathbf{z}}_i^0\rangle}{2},0,1
       \right),
    \label{eq:feature_drift}\\
    D_i^p
    &= \frac{1}{|\mathcal{C}_i|}
       \sum_{c\in\mathcal{C}_i}
       \left|\mu_{i,c}^{p,t}-\mu_{i,c}^{p,0}\right|.
    \label{eq:probability_drift}
\end{align}
For an LV-only exemplar, $\mathcal{C}_i=\{\mathrm{LV}\}$.

\paragraph{Boundary-response drift.}
We measure changes in the mean and variability of the auxiliary
boundary response:
\begin{equation}
    D_i^b
    = \min\!\left\{1,\,
      \frac{|\mu_i^{b,t}-\mu_i^{b,0}|
           +|\sigma_i^{b,t}-\sigma_i^{b,0}|}{2}\right\}.
    \label{eq:boundary_drift}
\end{equation}
These statistics summarize the distribution of boundary responses.
The structural term below measures where the predicted contours change.

\paragraph{Structural drift.}
Let $Y_{i,c}^a$ be the binary mask for class $c$ extracted from
$\hat{\mathbf{Y}}_i^a$, let $\partial Y$ denote its contour, and let
$\kappa(Y)$ count its 8-connected components. We compute contour
disagreement and component-count change as
\begin{align}
    E_{i,c}
    &= \frac{|\partial Y_{i,c}^0\triangle\partial Y_{i,c}^t|}
            {|\partial Y_{i,c}^0\cup\partial Y_{i,c}^t|+\epsilon},
    \label{eq:contour_drift}\\
    C_{i,c}
    &= \min\!\left\{1,\,
       \frac{|\kappa(Y_{i,c}^t)-\kappa(Y_{i,c}^0)|}{3}\right\},
    \label{eq:component_drift}
\end{align}
where $\triangle$ denotes symmetric difference. Let
$\mathcal{C}_i^*$ contain the classes in $\mathcal{C}_i$ that appear
in either the source or current prediction. For a nonempty
$\mathcal{C}_i^*$, structural drift is
\begin{equation}
    D_i^s
    = \frac{1}{|\mathcal{C}_i^*|}
      \sum_{c\in\mathcal{C}_i^*}
      \left(0.7E_{i,c}+0.3C_{i,c}\right).
    \label{eq:structure_drift}
\end{equation}
Excluding classes absent from both predictions prevents agreement on
empty masks from diluting changes in the represented anatomy.

\subsection{Drift-Conditioned Replay Weights}
\label{subsec:normalization}

We aggregate the four discrepancies into an exemplar-level score:
\begin{equation}
    D_i=0.30D_i^f+0.25D_i^p+0.20D_i^b+0.25D_i^s.
    \label{eq:unified_drift}
\end{equation}
The same bounded normalization is applied to the aggregate, boundary, and structural scores:
\begin{equation}
    \mathcal{N}(d)
    = \operatorname{clip}\!\left(
      \frac{d-0.05}{0.50-0.05},0,1\right).
    \label{eq:drift_normalization}
\end{equation}
Writing $\bar D_i=\mathcal{N}(D_i)$,
$\bar D_i^b=\mathcal{N}(D_i^b)$, and
$\bar D_i^s=\mathcal{N}(D_i^s)$, the branch weights are
\begin{align}
    b_i &= 1+0.75\left(0.5\bar D_i+0.5\bar D_i^b\right),
    \label{eq:boundary_multiplier}\\
    s_i &= 1+0.4\bar D_i+0.6\bar D_i^s.
    \label{eq:structure_multiplier}
\end{align}
Normalization gives $1\leq b_i\leq1.75$ and $1\leq s_i\leq2$.
The structural branch assigns more weight to changes in the predicted
mask, while the boundary branch gives equal weight to aggregate drift
and boundary-response drift.

\subsection{Segmentation and Anatomical Losses}
\label{subsec:overall_objective}

For a target fully labeled in the shared output space, the regional
loss combines multiclass cross-entropy and Dice loss. For an LV-only
target, binary cross-entropy and Dice loss are evaluated using the LV
probability and binary LV mask:
\begin{align}
    \mathcal{L}_{\mathrm{seg}}^{\mathrm{full}}
    &=\mathcal{L}_{\mathrm{CE}}
      +\mathcal{L}_{\mathrm{Dice}}^{\mathrm{multi}},
    \label{eq:full_loss}\\
    \mathcal{L}_{\mathrm{seg}}^{\mathrm{LV}}
    &=\mathcal{L}_{\mathrm{BCE}}^{\mathrm{LV}}
      +\mathcal{L}_{\mathrm{Dice}}^{\mathrm{LV}}.
    \label{eq:lv_loss}
\end{align}
For exemplar $i$, the boundary loss $\ell_i^b$ is pixel-averaged
binary cross-entropy between $\mathbf{Q}_i$ and the boundary target
extracted from its annotation. The base loss on a batch $\mathcal{B}$
is
\begin{equation}
    \mathcal{L}_{\mathrm{base}}(\mathcal{B})
    =\frac{1}{|\mathcal{B}|}\sum_{i\in\mathcal{B}}
     \left(\ell_i^{\mathrm{seg}}+\lambda_b\ell_i^b\right),
    \qquad \lambda_b=0.30,
    \label{eq:base_objective}
\end{equation}
where $\ell_i^{\mathrm{seg}}$ is evaluated according to the
exemplar's annotation mode.

For replay targets with MYO and LA supervision, the structural loss
uses a class-specific Tversky loss:
\begin{align}
    \ell_{i,c}^{\mathrm{Tv}}
    &=1-\frac{TP_{i,c}+\epsilon}
        {TP_{i,c}+0.30FP_{i,c}+0.70FN_{i,c}+\epsilon},
    \label{eq:tversky}\\
    \ell_i^s
    &=1.25\,\ell_{i,\mathrm{MYO}}^{\mathrm{Tv}}
      +\ell_{i,\mathrm{LA}}^{\mathrm{Tv}}.
    \label{eq:structural_sample_loss}
\end{align}
Here $TP$, $FP$, and $FN$ are soft counts computed from the class
probabilities and target masks. The larger false-negative coefficient
penalizes missed anatomy. For LV-only exemplars, $\ell_i^s=0$.

\subsection{Epoch-Level Replay Regulation and Training Objective}
\label{subsec:adaptive_replay}

Both the baseline and DCSRL adjust the global replay coefficient
$\lambda_r$ using validation Dice on earlier tasks. Let $A_j^*$ be the
best previously observed validation Dice for task $j$, and let
$A_j^{(e)}$ be its validation Dice after epoch $e$ of the current task.
For the nonempty set $\mathcal{P}_t$ of earlier tasks, the controller
computes
\begin{align}
    F^{(e)}
    &=\frac{1}{|\mathcal{P}_t|}
      \sum_{j\in\mathcal{P}_t}\max(0,A_j^*-A_j^{(e)}),
    \label{eq:mean_forgetting}\\
    \widetilde F^{(e)}
    &=0.70\widetilde F^{(e-1)}+0.30F^{(e)}.
    \label{eq:ema_forgetting}
\end{align}
The replay coefficient is $1.0$ during the first three epochs.
Following warm-up, the coefficient for the next epoch is
\begin{equation}
    \lambda_r^{(e+1)}
    =\operatorname{clip}\!\left(
      1.0+4.0\widetilde F^{(e)},\,0.75,\,1.75\right).
    \label{eq:adaptive_replay}
\end{equation}

Let $\mathcal{B}_c$ and $\mathcal{B}_r$ denote the current and replay
batches, respectively, and let $B_r=|\mathcal{B}_r|$. For $B_r>0$,
the complete objective is
\begin{equation}
\begin{aligned}
    \mathcal{L}_{\mathrm{total}}
    ={}&\mathcal{L}_{\mathrm{base}}(\mathcal{B}_c)
       +\lambda_r^{(e)}
        \mathcal{L}_{\mathrm{base}}(\mathcal{B}_r)\\
     &+\frac{1}{B_r}\sum_{i\in\mathcal{B}_r}
       \left[
         \lambda_b(b_i-1)\ell_i^b
         +\lambda_s s_i\ell_i^s
       \right],
    \qquad \lambda_s=0.30.
\end{aligned}
\label{eq:overall_dcsrl_objective}
\end{equation}
The boundary term uses $b_i-1$ because the base replay loss already
includes boundary supervision. The factor $s_i$ combines the
baseline structural term and its drift-dependent increment.
Setting $b_i=s_i=1$ for all replayed exemplars recovers the
reliability-guided replay objective. On the first task, training uses
only $\mathcal{L}_{\mathrm{base}}(\mathcal{B}_c)$. Reference storage
and drift estimation add training costs, while inference uses the
shared segmentation network without additional trainable parameters.

Algorithm~\ref{alg:dcsrl} summarizes the complete
training procedure of the proposed DCSRL framework.
The method follows a sequential learning strategy in
which current-task optimization is supplemented by
historical exemplars maintained through a
reliability-guided replay policy. For each replayed
exemplar, DCSRL measures feature, probability,
boundary, and structural drift relative to its frozen
source-task reference signature. The resulting drift
measurements determine exemplar-specific boundary
and structural loss weights, allowing anatomical
supervision to respond to changes in previously
learned representations and predictions. 
Drift measurements are computed without gradient
tracking and serve exclusively as exemplar-specific
supervision coefficients rather than differentiable
optimization targets. The segmentation network
operates in training mode during parameter updates. Following each task, the replay memory is updated according to the underlying selection and replacement policy, while newly retained exemplars are associated with frozen reference signatures for subsequent tasks.

\begin{algorithm}[!t]
\caption{Training Procedure of Drift-Conditioned
Structural Replay Loss (DCSRL)}
\label{alg:dcsrl}

\footnotesize
\begin{algorithmic}[1]

\Require Sequential datasets
$\{\mathcal{D}_t\}_{t=1}^{T}$;
segmentation model $f_{\theta}$;
memory capacity $M$;
drift weights $\mathbf{w}$

\Ensure Continually trained model $f_{\theta_T}$

\State Initialize model parameters $\theta$
\State Initialize replay memory $\mathcal{M}\gets\emptyset$

\For{$t=1,\ldots,T$}

    \State Initialize Adam optimizer and
    replay controller

    \State Initialize early-stopping counter

    \For{$e=1,\ldots,E_t$}

        \State Set $f_{\theta}$ to training mode

        \For{each current-task batch $\mathcal{B}_c$}

            \State Compute current-task loss
            $\mathcal{L}_c \gets
            \mathcal{L}_{\mathrm{base}}(\mathcal{B}_c)$

            \State $\mathcal{L}_{\mathrm{total}}
            \gets \mathcal{L}_c$

            \If{$\mathcal{M}\neq\emptyset$}

                \State Sample replay batch
                $\mathcal{B}_r\sim\mathcal{M}$
                using baseline priorities

                \State Compute base replay loss
                $\mathcal{L}_r\gets
                \mathcal{L}_{\mathrm{base}}(\mathcal{B}_r)$

                \State Compute baseline structural
                replay loss

                \Statex \textit{Source-referenced drift estimation:}

                \State Evaluate replay responses
                without gradient tracking

                \For{each exemplar $i\in\mathcal{B}_r$}

                    \State Retrieve frozen
                    source reference $\mathcal{R}_i$

                    \State Compute feature and
                    probability drift $D_i^f,D_i^p$

                    \State Compute boundary and
                    structural drift $D_i^b,D_i^s$

                    \State Aggregate drift
                    $D_i\gets
                    w_fD_i^f+w_pD_i^p+
                    w_bD_i^b+w_sD_i^s$

                    \State Normalize
                    $D_i,D_i^b,D_i^s$
                    using Eq.~\eqref{eq:drift_normalization}

                    \State Compute multipliers
                    $b_i,s_i$ using
                    Eqs.~\eqref{eq:boundary_multiplier}
                    --\eqref{eq:structure_multiplier}

                \EndFor

                \State Restore training mode

                \State Compute
                $\mathcal{L}_{\mathrm{total}}$
                using Eq.~\eqref{eq:overall_dcsrl_objective}

            \EndIf

            \State Update $\theta$ using Adam
            and mixed-precision training

        \EndFor

        \State Evaluate all previously learned
        tasks and the current task

        \State Update replay coefficient
        $\lambda_r$ for the next epoch using
        Eqs.~\eqref{eq:mean_forgetting}
        --\eqref{eq:adaptive_replay}

        \State Update early-stopping counter
        using current-task validation Dice

        \If{early-stopping patience is reached}
            \State \textbf{break}
        \EndIf

    \EndFor

    \State Evaluate and save the
    end-of-task model checkpoint

    \State Score current-task training exemplars
    using reliability-guided priorities

    \State Construct task-specific candidate memory

    \State Compute frozen source-task
    reference signatures for candidates

    \State Merge candidates into $\mathcal{M}$
    using the baseline allocation
    and replacement policy

    \State Retain reference signatures
    for admitted exemplars

\EndFor

\State \Return $f_{\theta_T}$

\end{algorithmic}
\end{algorithm}

\section{Experimental Setup}
\label{sec:experimental_setup}

\subsection{Datasets and Continual-Learning Protocol}
\label{subsec:datasets}
\label{subsec:cl_protocol}

All methods followed the sequence
\begin{equation}
    \mathrm{CAMUS}
    \rightarrow \mathrm{CardiacNet}
    \rightarrow \mathrm{EchoNet\text{-}Dynamic}.
    \label{eq:experimental_sequence}
\end{equation}
At each stage, the model trained on the current dataset. Replay-based
methods could also access a bounded memory of earlier training
examples. After each stage, we evaluated the model across tasks to
measure final performance, retention, and generalization to datasets
encountered later in the sequence. 

\paragraph{CAMUS.}
The CAMUS dataset \cite{leclerc2019deep} contains two-dimensional echocardiographic examinations from 500 patients. Each examination includes apical two-chamber (2CH) and four-chamber (4CH) views, with expert annotations provided at end-diastole (ED) and end-systole (ES). The dataset contains patients with different image qualities and cardiac conditions, making it a useful benchmark for multi-structure echocardiographic segmentation. We followed the official patient-level partition: 400 patients were used for training, 50 for validation, and the remaining 50 for testing. Keeping the partition at the patient level ensures that images originating from the same patient cannot appear across training and evaluation subsets.

For our experiments, the segmentation label space was standardized to four classes: BG, LV, MYO, and LA, represented by labels 0, 1, 2, and 3, respectively. Only the annotated ED and ES acquisitions were used. Echocardiographic images and their corresponding masks were resized to $256\times256$ pixels. Image intensities were normalized to the range $[0,1]$, while nearest-neighbor interpolation was used for segmentation masks to preserve discrete anatomical labels.

\subsubsection{CardiacNet}
\label{subsubsec:cardiacnet}

CardiacNet was introduced by Yang et al.~\cite{yang2024cardiacnet} for
echocardiographic cardiac-disease assessment. Its inclusion as the second
task introduces a distinct echocardiographic distribution after CAMUS rather
than another partition of the same source dataset.

For the present segmentation experiments, the available NIfTI image-label
pairs were processed slice-wise and harmonized with the common segmentation
space

\begin{equation}
\mathcal{C}_{\mathrm{CardiacNet}}
=
\{
\mathrm{BG},
\mathrm{LV},
\mathrm{MYO},
\mathrm{LA}
\}.
\end{equation}

Following the label convention used throughout the experimental pipeline,
CardiacNet label 4 was mapped to background,

\begin{equation}
4\longrightarrow0.
\label{eq:cardiacnet_mapping}
\end{equation}

Slices containing no foreground cardiac structure after label harmonization
were excluded from training. CardiacNet therefore evaluates whether the model
can adapt to a second echocardiographic distribution while retaining the
multiclass anatomical representation acquired from CAMUS.


\paragraph{EchoNet-Dynamic.}
EchoNet-Dynamic \cite{ouyang2020video} is a large-scale echocardiographic video dataset containing 10,030 apical four-chamber videos acquired during routine clinical practice. In contrast to CAMUS and CardiacNet, EchoNet-Dynamic provides expert LV tracings rather than complete LV-MYO-LA segmentation annotations.  Expert contour coordinates were converted into binary LV masks, and the corresponding annotated frames were extracted from the original videos. Frames and masks were subsequently resized to $256\times256$ pixels. Because MYO and LA GT annotations are unavailable in EchoNet-Dynamic, these structures were not treated as background during optimization. Instead, EchoNet-Dynamic samples were handled using LV-only partial supervision. Evaluation on this dataset was likewise restricted to the LV. This distinction is particularly relevant to the CL setting: the final task changes not only the image domain but also the available anatomical supervision. For EchoNet-Dynamic, we follow the original dataset partition.

\subsection{Continual-Learning Protocol}

All methods followed the task sequence in
Eq.~\eqref{eq:experimental_sequence}. Let $A_{i,j}$ denote the Dice
performance on task $j$ after completing training on task $i$. 
At stage $t$, training was performed primarily on the current dataset
$\mathcal{D}_t$. Replay-based methods were allowed to access only their
bounded replay memory from previous tasks; complete historical datasets were not revisited. This constraint distinguishes the experimental protocol from joint retraining and ensures that all replay-based methods operate under a common historical-data budget.

For the CAMUS dataset, training and validation subsets are established using either the official data partitioning protocol or an 80\%–20\% patient-level split to prevent data leakage. Similarly, CardiacNet is divided into internal training and validation subsets when used for model development, whereas the entire dataset serves as an independent external test set for assessing cross-dataset generalization.

EchoNet-Dynamic retained its predefined TRAIN and VAL partitions. The entire continual sequence was independently repeated for

\begin{equation}
\mathcal{S}=\{42,123,2025\}.
\end{equation}

For all replay-based comparisons, the total replay-memory capacity was fixed
at

\begin{equation}
M=1800
\end{equation}

examples. The current-task mini-batch size was 16 and the replay mini-batch
size was 8. Maintaining a common memory budget and replay batch size prevents
performance differences from being attributed simply to unequal access to
historical observations.

\subsection{Preprocessing and Training}
\label{subsec:preprocessing}
\label{subsec:implementation}

We converted images to grayscale, resized them to $256\times256$ using
area interpolation, and applied per-image min--max normalization:
\begin{equation}
    x'=\frac{x-\min(x)}{\max(x)-\min(x)+\epsilon}.
    \label{eq:minmax}
\end{equation}
Masks were resized using nearest-neighbor interpolation. Training
augmentations comprised horizontal flipping and rotation by multiples
of $90^\circ$, each applied with probability 0.5. Each image and its
mask received the same geometric transformation. Boundary targets were
derived from transitions between neighboring mask labels.

All methods used the U-Net backbone described in
Section~\ref{sec:method} \cite{ronneberger2015u}. The four encoder
stages had 32, 64, 128, and 256 channels, followed by a 512-channel
bottleneck. Each block contained two $3\times3$ convolutions with batch
normalization and ReLU activation. The encoder used $2\times2$ max
pooling, and the decoder used transposed convolutions and skip
connections. The final decoder features supplied a four-channel
segmentation head and a single-channel boundary head. DCSRL extracted
its reference features by global average pooling of the bottleneck.

\begin{table}[!t]
\centering
\caption{Training hyperparameters and configuration of the
proposed DCSRL framework.}
\label{tab:training_hyperparameters}

\footnotesize
\setlength{\tabcolsep}{4pt}
\renewcommand{\arraystretch}{1.08}

\begin{tabularx}{\linewidth}{
    >{\raggedright\arraybackslash}X
    >{\raggedright\arraybackslash}p{2.9cm}
}
\toprule
\textbf{Hyperparameter} & \textbf{Value} \\
\midrule

Segmentation backbone & U-Net with boundary head \\
Input resolution & $256 \times 256$ \\
Input channels & 1 (grayscale) \\
Output classes & 4 \\
Initial feature channels & 32 \\
Optimizer & Adam \\
Initial learning rate & $1 \times 10^{-4}$ \\
Weight decay & $1 \times 10^{-5}$ \\
Maximum epochs per task & 20 \\
Early stopping patience & 8 epochs \\
Current-task batch size & 16 \\
Replay batch size & 8 \\
Mixed-precision training & Enabled \\
Random seeds & 42, 123, 2025 \\
\midrule

Replay memory capacity & 1800 exemplars \\
Memory allocation (CAMUS/CardiacNet/EchoNet)
& 25\% / 50\% / 25\% \\
Initial replay coefficient $\lambda_r$ & 1.0 \\
Adaptive replay range & $[0.75,\,1.75]$ \\
Replay-controller gain & 4.0 \\
Replay-controller EMA & 0.70 \\
Replay warm-up & 3 epochs \\
Boundary-loss weight $\lambda_b$ & 0.30 \\
Structural replay weight $\lambda_s$ & 0.30 \\
MYO structural weight & 1.25 \\
LA structural weight & 1.00 \\
Tversky coefficients $(\alpha,\beta)$ & $(0.30,\,0.70)$ \\
\midrule

Drift weights (feature/probability/
boundary/structure) & $0.30/0.25/0.20/0.25$ \\
Drift normalization bounds & $[0.05,\,0.50]$ \\
Boundary-conditioning gain & 0.75 \\
Structural-conditioning gain & 1.00 \\
Maximum boundary multiplier & 1.75 \\
Maximum structural multiplier & 2.00 \\
Numerical stabilizer $\epsilon$ & $10^{-6}$ \\

\bottomrule
\end{tabularx}

\par\smallskip
\begin{minipage}{\linewidth}
\footnotesize
\textit{Note:} EMA denotes exponential moving average.
The three random seeds are 42, 123, and 2025.
Replay-memory allocation percentages correspond
to CAMUS, CardiacNet, and EchoNet-Dynamic,
respectively.
\end{minipage}

\end{table}

As illustrated in Table \ref{tab:training_hyperparameters}, we trained with Adam at an initial learning rate of $10^{-4}$ and
weight decay of $10^{-5}$. Each task ran for up to 20 epochs with
automatic mixed precision and early stopping after eight epochs without
validation improvement. Current-task batches contained 16 images.
Replay-based methods used a capacity of 1800 exemplars and a replay
batch size of eight. This budget fixes the number of retained examples;
the storage required per exemplar depends on the method. The regional
objective followed the annotation mode defined in
Section~\ref{sec:method}, with boundary-loss weight $\lambda_b=0.30$.

\subsection{Comparison Methods}
\label{subsec:comparison_methods}

We compared DCSRL with 11 baselines using the same task order,
backbone, preprocessing, and evaluation protocol.

\paragraph{Fine-tuning, regularization, and distillation.}
Fine-tuning optimized the network sequentially on each current task.
Elastic Weight Consolidation (EWC) used a diagonal Fisher estimate
computed from up to 40 training batches per task and a regularization
coefficient of 20 \cite{kirkpatrick2017overcoming}. Learning without
Forgetting (LwF) used the preceding model as a frozen teacher, with
distillation weight 0.5 and temperature 2 \cite{li2017learning}.
Distillation used four-channel predictions for multiclass targets and
LV-versus-non-LV predictions for EchoNet-Dynamic.

\paragraph{Memory construction and retrieval.}
Plain Replay used first-in-first-out replacement and uniform retrieval.
Random Replay used reservoir sampling for admission and uniform
retrieval \cite{chaudhry2019tiny}. The iCaRL-style baseline used
feature herding within class-containing subsets, followed by global
herding to fill the remaining memory \cite{rebuffi2017icarl}. Maximally
Interfered Retrieval (MIR) prioritized stored examples expected to incur
a larger loss after a current-task update \cite{aljundi2019online}.
We also included an AMSS-inspired exemplar-selection baseline adapted
from continual semantic segmentation \cite{zhu2023continual}.

\paragraph{Historical-output replay.}
Dark Experience Replay (DER) constrained current predictions using
logits stored with historical examples. DER++ additionally applied
supervised replay to the stored labels \cite{buzzega2020dark}.

\paragraph{Reliability-guided replay.}
We used the configuration defined in
Section~\ref{subsubsec:reliability_guided_replay} as the direct
control for DCSRL. Both configurations share the exemplar priorities,
replacement and sampling rules, epoch-level replay controller, and
base structural loss. Their comparison evaluates the additional
effect of exemplar-specific drift conditioning.

\subsection{Evaluation Metrics}
\label{subsec:evaluation_metrics}

\subsubsection{Segmentation Performance}
\label{subsubsec:segmentation_metrics}

After the final task, we evaluated regional overlap using Dice and
intersection over union (IoU). For a predicted region $P$ and
reference region $G$,
\begin{align}
    \mathrm{Dice}(P,G)
    &=\frac{2|P\cap G|}{|P|+|G|},
    \label{eq:dice_metric}\\
    \mathrm{IoU}(P,G)
    &=\frac{|P\cap G|}{|P\cup G|}.
    \label{eq:iou_metric}
\end{align}
CAMUS scores were averaged over LV, MYO, and LA. CardiacNet scores
were averaged over the three foreground channels retained by its
experimental encoding. EchoNet-Dynamic was scored on LV alone.

For surface evaluation, we computed the distance from each predicted
boundary point to its nearest reference boundary point and repeated
the calculation in the reverse direction. The 95th-percentile
Hausdorff distance (HD95) and average surface distance (ASD) were the
95th percentile and mean, respectively, of the pooled bidirectional
distances. Boundary F1 was the harmonic mean of boundary precision and
recall, using a matching tolerance of two pixels in each direction.
Higher Dice, IoU, and boundary F1 indicate better agreement. Lower
HD95 and ASD indicate smaller surface errors.

\subsubsection{Continual-Learning Performance}
\label{subsubsec:continual_metrics}

Let $A_{i,j}$ denote the Dice score on task $j$ after training on
task $i$, and let $T=3$. We report final average accuracy (FAA),
backward transfer (BWT), and forward transfer (FWT). In this study,
FAA is the unweighted mean final Dice across datasets. BWT measures
the change on earlier tasks relative to their scores immediately
after training \cite{lopez2017gradient}:
\begin{align}
    \mathrm{FAA}
    &=\frac{1}{T}\sum_{j=1}^{T}A_{T,j},
    \label{eq:faa}\\
    \mathrm{BWT}
    &=\frac{1}{T-1}\sum_{j=1}^{T-1}
      \left(A_{T,j}-A_{j,j}\right).
    \label{eq:bwt}
\end{align}
A positive BWT indicates that subsequent training improved the mean
Dice on earlier tasks.

Our FWT measure is the mean Dice on future tasks before the model
receives training data from those tasks:
\begin{equation}
    \mathrm{FWT}
    =\frac{1}{|\Omega|}\sum_{(i,j)\in\Omega}A_{i,j},
    \qquad
    \Omega=\{(i,j):i<j,\ A_{i,j}\text{ is available}\}.
    \label{eq:fwt}
\end{equation}
This definition reports absolute pre-training performance on future
tasks. Its numerical scale differs from FWT definitions that subtract
a random-initialization reference.

\subsection{Ablation and Computational Cost}
\label{subsec:ablation_protocol}

We evaluated six ablations under the full model's training and memory
settings. Four variants individually removed feature, probability,
boundary, or structural drift. Two further variants disabled boundary
conditioning or structural conditioning while retaining the underlying
supervised loss. We evaluated each variant on all three datasets using
the five segmentation metrics. Reliability-guided replay provided the
control with both conditioning branches disabled in the main comparison.

For DCSRL and reliability-guided replay, we recorded trainable parameter
count, per-task training time, peak GPU memory, and persistent replay
storage after each task. DCSRL storage included its source-reference
signatures. For each task, training overhead was calculated as the
percentage increase in mean training time relative to reliability-guided
replay.

\subsection{Statistical Analysis}
\label{subsec:statistical_analysis}

We repeated each principal experiment with seeds 42, 123, and 2025 and
report the mean and sample standard deviation across runs. For the
direct comparison, we paired DCSRL and reliability-guided replay by
seed. Let $d_k=x_k^{\mathrm{DCSRL}}-x_k^{\mathrm{baseline}}$ denote
the difference for seed $k$. We used a two-sided paired $t$-test and
calculated the paired standardized effect size
$d_z=\bar d/s_d$, where $\bar d$ and $s_d$ are the mean and sample
standard deviation of the paired differences. Negative differences
favor DCSRL for HD95 and ASD, and positive differences favor DCSRL
for the overlap and boundary F1 scores. With three seed pairs, the
analysis emphasizes the magnitude and direction of the observed
differences. The reported standard deviations describe variation
across training runs.

\section{Experimental Results}
\label{sec:results}

\subsection{Final Performance and Knowledge Transfer}
\label{subsec:overall_cl_results}

DCSRL achieved the highest final average accuracy (FAA) across the
three seeds, with a mean final Dice of $0.8784\pm0.0017$, compared
with $0.8718\pm0.0029$ for reliability-guided replay (Table \ref{tab:continual_learning_comparison}), the next-best
method. This represents
an absolute gain of 0.66 Dice percentage points under the same
replay-memory policy.

\begin{table*}[t]
\centering

\caption{Continual-learning performance
of the compared methods.}
\label{tab:continual_learning_comparison}

\setlength{\tabcolsep}{8pt}
\renewcommand{\arraystretch}{1.15}

\begin{tabular}{lccc}
\toprule

\textbf{Method}
& \textbf{FAA $\uparrow$}
& \textbf{BWT $\uparrow$}
& \textbf{FWT $\uparrow$} \\

\midrule

Fine-tuning
& $0.4733 \pm 0.0130$
& $-0.6152 \pm 0.0219$
& $0.3100 \pm 0.0469$ \\

EWC
& $0.5269 \pm 0.0078$
& $-0.5369 \pm 0.0144$
& $0.3100 \pm 0.1011$ \\

LwF
& $0.3408 \pm 0.0291$
& $-0.2639 \pm 0.0183$
& $0.3416 \pm 0.0494$ \\

Plain Replay
& $0.6758 \pm 0.0280$
& $-0.2606 \pm 0.0423$
& $0.3853 \pm 0.0277$ \\

Random Replay
& $0.8495 \pm 0.0079$
& $-0.0019 \pm 0.0104$
& $0.3853 \pm 0.0277$ \\

iCaRL Replay
& $0.8397 \pm 0.0108$
& $-0.0167 \pm 0.0231$
& $0.3853 \pm 0.0277$ \\

\midrule

DER
& $0.7934 \pm 0.0378$
& $-0.0282 \pm 0.0018$
& $0.4173 \pm 0.0070$ \\

DER++
& $0.8496 \pm 0.0062$
& $\mathbf{0.0546 \pm 0.0313}$
& $0.4265 \pm 0.0177$ \\

MIR
& $0.8534 \pm 0.0048$
& $0.0243 \pm 0.0267$
& $0.3998 \pm 0.0428$ \\

AMSS-inspired
& $0.8521 \pm 0.0024$
& $0.0390 \pm 0.0407$
& $\mathbf{0.4440 \pm 0.0221}$ \\

\midrule

Reliability-Guided Replay
& $0.8718 \pm 0.0029$
& $0.0489 \pm 0.0116$
& $0.4366 \pm 0.0357$ \\

\midrule

\textbf{DCSRL (Proposed)}
& $\mathbf{0.8784 \pm 0.0017}$
& $0.0003 \pm 0.0049$
& $0.4080 \pm 0.0299$ \\

\bottomrule
\end{tabular}

\par\smallskip
\begin{minipage}{\linewidth}
\footnotesize
\textit{Note:} Values are reported as mean
$\pm$ standard deviation over three random
seeds (42, 123, and 2025).
Higher FAA, BWT, and FWT indicate better
performance. Bold indicates the best mean
for each metric.
\end{minipage}

\end{table*}

The transfer measures showed a different pattern. DCSRL's backward
transfer was close to zero ($0.0003\pm0.0049$), indicating little
average change on earlier tasks relative to their scores immediately
after training. Reliability-guided replay achieved higher backward
transfer ($0.0489\pm0.0116$) and higher forward transfer
($0.4366\pm0.0357$ versus $0.4080\pm0.0299$). Among all methods,
DER++ achieved the highest backward transfer ($0.0546$), and the
AMSS-inspired method achieved the highest forward transfer ($0.4440$).

\subsection{Final Segmentation Performance}
\label{subsec:final_segmentation_results}

After the full training sequence, DCSRL improved the mean result in
14 of 15 dataset--metric combinations relative to reliability-guided
replay (Table~\ref{tab:final_segmentation_all_methods}). CardiacNet
showed the largest Dice gain, from $0.8096$ to $0.8218$, and the
largest reductions in surface error. HD95 decreased from $16.0261$
to $10.2148$ ($36.3\%$), and ASD decreased from $5.2089$ to $3.7972$
($27.1\%$). Boundary F1 also increased from $0.5765$ to $0.5998$.

\begin{table*}[t]
\centering
\caption{Final segmentation performance
after sequential echocardiographic training.}
\label{tab:final_segmentation_all_methods}

\setlength{\tabcolsep}{1.5pt}
\renewcommand{\arraystretch}{1.05}
\tiny

\begin{tabular}{@{}llccccc@{}}
\toprule
\textbf{Dataset}
& \textbf{Method}
& \textbf{Dice $\uparrow$}
& \textbf{IoU $\uparrow$}
& \textbf{HD95 $\downarrow$}
& \textbf{ASD $\downarrow$}
& \textbf{Boundary F1 $\uparrow$} \\
\midrule


\multirow{11}{*}{CAMUS}
& Fine-tuning
& $0.3268 \pm 0.0289$
& $0.2608 \pm 0.0293$
& $77.9952 \pm 27.9890$
& $38.0120 \pm 17.1613$
& $0.1100 \pm 0.0195$ \\

& EWC
& $0.4564 \pm 0.0202$
& $0.3610 \pm 0.0254$
& $31.8637 \pm 4.3128$
& $10.6272 \pm 2.2729$
& $0.1690 \pm 0.0440$ \\

& LwF
& $0.3767 \pm 0.0269$
& $0.3073 \pm 0.0186$
& $53.1436 \pm 7.0249$
& $19.4207 \pm 5.7180$
& $0.1728 \pm 0.0392$ \\

& Plain Replay
& $0.3392 \pm 0.0745$
& $0.2472 \pm 0.0705$
& $41.7560 \pm 7.3061$
& $17.4481 \pm 3.4357$
& $0.1417 \pm 0.0385$ \\

& Random Replay
& $0.8870 \pm 0.0032$
& $0.7986 \pm 0.0051$
& $8.3637 \pm 0.4174$
& $3.1037 \pm 0.0779$
& $0.5522 \pm 0.0125$ \\

& iCaRL Replay
& $0.8810 \pm 0.0067$
& $0.7891 \pm 0.0105$
& $9.1623 \pm 0.6260$
& $3.3030 \pm 0.1932$
& $0.5463 \pm 0.0073$ \\

\addlinespace[2pt]

& DER
& $0.8760 \pm 0.0017$
& $0.7814 \pm 0.0030$
& $21.1039 \pm 21.3896$
& $5.6569 \pm 4.1085$
& $0.5069 \pm 0.0268$ \\

& DER++
& $0.8882 \pm 0.0005$
& $0.8005 \pm 0.0008$
& $\mathbf{7.9559 \pm 0.2681}$
& $2.9995 \pm 0.0437$
& $0.5574 \pm 0.0039$ \\

& MIR
& $0.8880 \pm 0.0059$
& $0.8002 \pm 0.0095$
& $8.6204 \pm 0.3489$
& $3.0953 \pm 0.1001$
& $0.5525 \pm 0.0078$ \\

& AMSS-inspired
& $0.8910 \pm 0.0052$
& $0.8049 \pm 0.0084$
& $8.1574 \pm 0.6152$
& $2.9944 \pm 0.2062$
& $0.5669 \pm 0.0170$ \\

& Reliability-Guided Replay 
& $0.8977 \pm 0.0024$
& $0.8114 \pm 0.0044$
&$8.2862 \pm 0.5435$
&$2.9359 \pm 0.1164$
&$0.5750 \pm 0.0096$

\\

\addlinespace[2pt]

& \textbf{DCSRL (Proposed)}
& $\mathbf{0.8986 \pm 0.0030}$
& $\mathbf{0.8172 \pm 0.0049}$
& $8.4169 \pm 0.7366$
& $\mathbf{2.8752 \pm 0.1583}$
& $\mathbf{0.5900 \pm 0.0033}$ \\

\midrule


\multirow{11}{*}{CardiacNet}
& Fine-tuning
& $0.1823 \pm 0.0108$
& $0.1205 \pm 0.0101$
& $102.0346 \pm 2.2724$
& $59.2795 \pm 2.4060$
& $0.1253 \pm 0.0114$ \\

& EWC
& $0.2104 \pm 0.0078$
& $0.1364 \pm 0.0055$
& $71.8151 \pm 3.4866$
& $36.4733 \pm 8.2664$
& $0.1202 \pm 0.0195$ \\

& LwF
& $0.1242 \pm 0.0118$
& $0.0719 \pm 0.0072$
& $87.4117 \pm 15.4822$
& $39.5617 \pm 9.8568$
& $0.0972 \pm 0.0100$ \\

& Plain Replay
& $0.7820 \pm 0.0195$
& $0.6439 \pm 0.0259$
& $20.9466 \pm 6.9804$
& $6.8668 \pm 2.0048$
& $0.5463 \pm 0.0433$ \\

& Random Replay
& $0.7516 \pm 0.0259$
& $0.6053 \pm 0.0321$
& $21.5832 \pm 4.2886$
& $7.3905 \pm 1.5881$
& $0.5146 \pm 0.0299$ \\

& iCaRL Replay
& $0.7280 \pm 0.0349$
& $0.5753 \pm 0.0413$
& $23.2788 \pm 6.1884$
& $7.8652 \pm 2.0925$
& $0.4850 \pm 0.0480$ \\

\addlinespace[2pt]

& DER
& $0.6021 \pm 0.1064$
& $0.4447 \pm 0.1071$
& $49.6329 \pm 27.5366$
& $19.0059 \pm 11.6101$
& $0.3144 \pm 0.0887$ \\

& DER++
& $0.7519 \pm 0.0196$
& $0.6041 \pm 0.0258$
& $22.3016 \pm 4.5647$
& $7.5371 \pm 1.1471$
& $0.4882 \pm 0.0186$ \\

& MIR
& $0.7646 \pm 0.0186$
& $0.6206 \pm 0.0239$
& $18.8856 \pm 4.3442$
& $6.5883 \pm 1.3403$
& $0.5252 \pm 0.0324$ \\

& AMSS-inspired
& $0.7553 \pm 0.0131$
& $0.6089 \pm 0.0172$
& $21.3780 \pm 2.8054$
& $7.3317 \pm 0.6696$
& $0.4968 \pm 0.0191$ \\

& Reliability-Guided Replay 
& $0.8096 \pm 0.0055$
& $0.6777 \pm 0.0106$
&$16.0261 \pm 0.9857$
&$5.2089 \pm 0.1841$
&$0.5765 \pm 0.0122$

\\

\addlinespace[2pt]

& \textbf{DCSRL (Proposed)}
& $\mathbf{0.8218 \pm 0.0029}$
& $\mathbf{0.6978 \pm 0.0041}$
& $\mathbf{10.2148 \pm 0.4996}$
& $\mathbf{3.7972 \pm 0.1541}$
& $\mathbf{0.5998 \pm 0.0071}$ \\

\midrule


\multirow{11}{*}{EchoNet}
& Fine-tuning
& $0.9109 \pm 0.0006$
& $0.8363 \pm 0.0010$
& $15.8577 \pm 0.4639$
& $4.3225 \pm 0.0816$
& $0.5137 \pm 0.0017$ \\

& EWC
& $0.9139 \pm 0.0005$
& $0.8414 \pm 0.0009$
& $15.4920 \pm 0.4814$
& $4.2132 \pm 0.0819$
& $0.5229 \pm 0.0041$ \\

& LwF
& $0.5215 \pm 0.0935$
& $0.3562 \pm 0.0830$
& $47.9719 \pm 4.1903$
& $15.5658 \pm 2.2557$
& $0.1662 \pm 0.0256$ \\

& Plain Replay
& $0.9061 \pm 0.0014$
& $0.8283 \pm 0.0023$
& $16.0567 \pm 0.1107$
& $4.4771 \pm 0.0612$
& $0.4927 \pm 0.0083$ \\

& Random Replay
& $0.9099 \pm 0.0009$
& $0.8347 \pm 0.0015$
& $15.8187 \pm 0.1701$
& $4.3632 \pm 0.0297$
& $0.5039 \pm 0.0011$ \\

& iCaRL Replay
& $0.9102 \pm 0.0007$
& $0.8352 \pm 0.0011$
& $15.6754 \pm 0.1058$
& $4.3406 \pm 0.0512$
& $0.5037 \pm 0.0063$ \\

\addlinespace[2pt]

& DER
& $0.9019 \pm 0.0072$
& $0.8213 \pm 0.0119$
& $41.9954 \pm 44.3794$
& $9.3054 \pm 8.2795$
& $0.4606 \pm 0.0626$ \\

& DER++
& $0.9085 \pm 0.0004$
& $0.8324 \pm 0.0008$
& $15.7607 \pm 0.0759$
& $4.3708 \pm 0.0147$
& $0.5026 \pm 0.0022$ \\

& MIR
& $0.9078 \pm 0.0014$
& $0.8311 \pm 0.0023$
& $15.8751 \pm 0.3035$
& $4.3957 \pm 0.0688$
& $0.5060 \pm 0.0008$ \\

& AMSS-inspired
& $0.9101 \pm 0.0016$
& $0.8350 \pm 0.0027$
& $15.9973 \pm 0.1409$
& $4.3776 \pm 0.0501$
& $0.5076 \pm 0.0052$ \\

\addlinespace[2pt]

& Reliability-Guided Replay 
& $0.9081 \pm 0.0023$
& $0.8331 \pm 0.0026$
&$15.3991 \pm 0.0927$
&$4.2779 \pm 0.0449$
&$0.5094 \pm 0.0042$

\\

& \textbf{DCSRL (Proposed)}
& $\mathbf{0.9148 \pm 0.0005}$
& $\mathbf{0.8429 \pm 0.0009}$
& $\mathbf{15.3023 \pm 0.2800}$
& $\mathbf{4.1372 \pm 0.0527}$
& $\mathbf{0.5283 \pm 0.0085}$ \\

\bottomrule
\end{tabular}

\par\smallskip
\begin{minipage}{\linewidth}
\footnotesize
\textit{Note:} Methods were trained sequentially
on CAMUS, CardiacNet, and EchoNet-Dynamic.
Values are reported as mean $\pm$ standard
deviation over seeds 42, 123, and 2025.
Higher Dice, IoU, and Boundary F1 are better,
whereas lower HD95 and ASD indicate better
performance. Bold indicates the best mean
within each dataset and metric.
\end{minipage}

\end{table*}

On CAMUS, Dice changed only slightly, from $0.8977$ to $0.8986$,
while boundary F1 increased from $0.5750$ to $0.5900$. CAMUS HD95
was the sole metric on which DCSRL performed worse than
reliability-guided replay, increasing from $8.2862$ to $8.4169$.
On EchoNet-Dynamic, DCSRL achieved the best mean value among the
compared methods on all five metrics, including Dice
($0.9148\pm0.0005$) and boundary F1 ($0.5283\pm0.0085$).

\subsection{Component Ablation}

The clearest ablation result was on EchoNet-Dynamic as illustrated in Table \ref{tab:dcsrl_ablation}.
Removing any one of the four drift signals or either conditioning
branch reduced performance on all five metrics. Across the six
ablations, boundary F1 ranged from $0.5131$ to $0.5160$, compared
with $0.5283$ for full DCSRL.

On CardiacNet, the full model achieved the best mean Dice, IoU, HD95,
and ASD. Removing structural drift yielded a marginally higher
boundary F1 ($0.6000$ versus $0.5998$). On CAMUS, removing
probability drift reduced HD95 from $8.4169$ to $8.1014$, and
disabling boundary conditioning reduced ASD from $2.8752$ to
$2.8307$. These differences across the three-seed means show that
the effect of conditioning varied by dataset and metric.

\subsection{Computational and Storage Costs}
\label{subsec:computational_results}

DCSRL and reliability-guided replay each contained 7,762,597
trainable parameters. DCSRL increased mean training time
by $2.82\%$ on CAMUS, $1.13\%$ on CardiacNet, and $15.36\%$ on
EchoNet-Dynamic. After the final task, storing the source-reference
signatures increased the replay footprint from $1{,}350.0$ to
$1{,}464.3$~MB, an additional $114.3$~MB ($8.47\%$).

\subsection{Qualitative Results}
\label{subsec:qualitative_results}

\begin{figure}[!t]
    \centering
    \includegraphics[width=\textwidth]{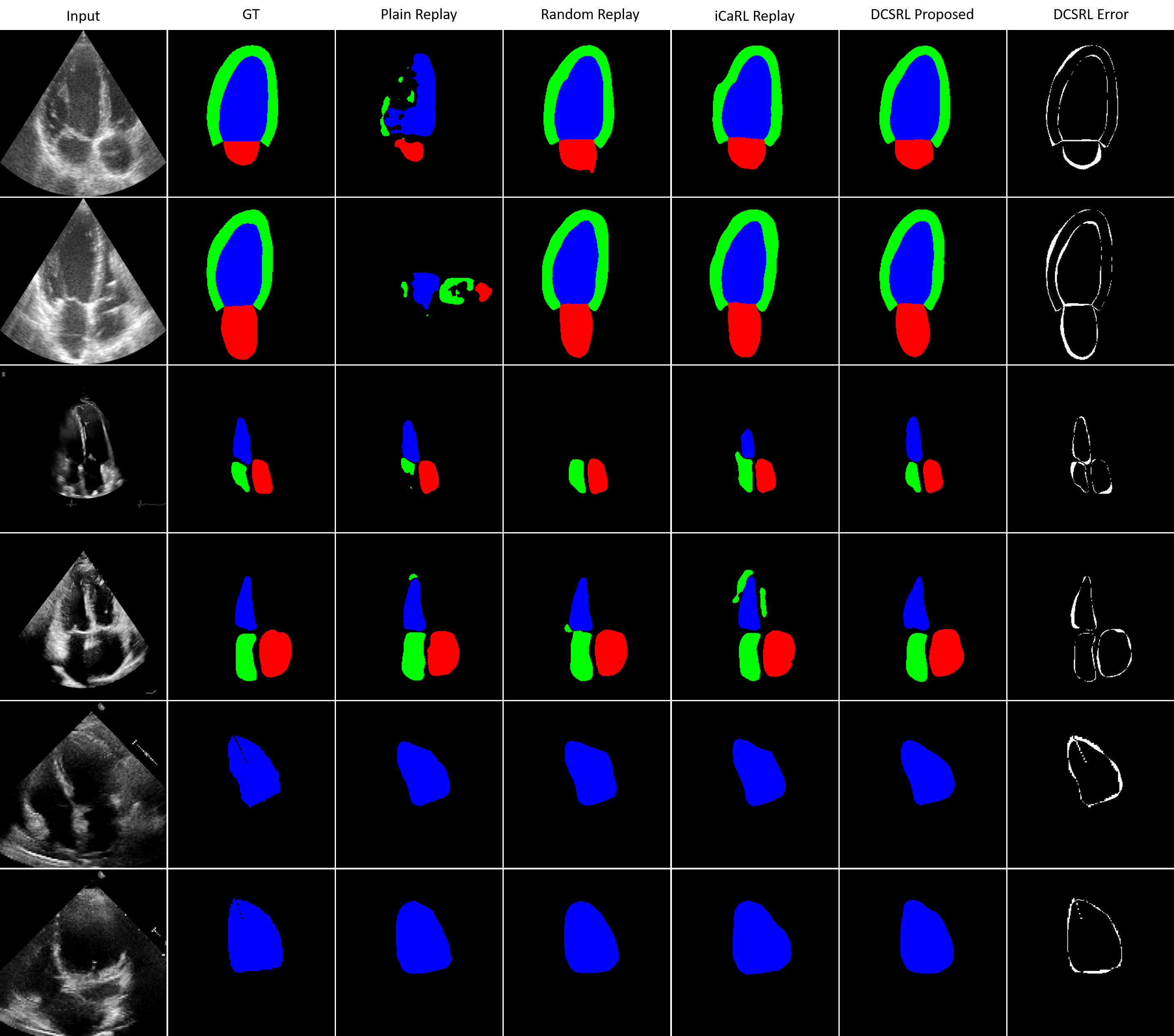}
    \caption{Qualitative comparison of final segmentation predictions
after continual echocardiographic training.
Two examples from each dataset are shown in order:
CAMUS, CardiacNet, and EchoNet-Dynamic.
Columns display the input echocardiogram, ground-truth
mask, predictions from Plain Replay, Random Replay,
iCaRL-style replay, and DCSRL, followed by the
DCSRL boundary-error map.
White pixels in the final column indicate
disagreement between predicted and reference contours.}
    \label{fig:qualitative_results}
\end{figure}

Figure~\ref{fig:qualitative_results} shows a recurring pattern across
the selected cases. In the multiclass examples, the predictions differ
in whether they recover each labeled region and preserve its position
relative to neighboring structures. In the LV-only examples, all
methods recover the ventricular cavity, and the visible differences
concern its outline.

On CAMUS, Plain Replay fragments the myocardial wall in the upper
example and produces disconnected regions in the lower example.
Random Replay, iCaRL-style replay, and DCSRL each recover a continuous
myocardial band around the ventricular cavity, with the atrium at its
base. On CardiacNet, Random Replay omits the upper blue region in the
first example, while iCaRL-style replay adds green fragments alongside
that region in the second. DCSRL recovers the upper region in both
cases and preserves its position relative to the two lower regions.
Its contour still differs locally from the reference, particularly
along the left edge of the upper region in the second example.

On EchoNet-Dynamic, the methods produce similar LV masks. The DCSRL
error maps show residual disagreement along the sloping ventricular
contour in the first example and near the apex and base in the second.
In both cases, the cavity remains a single connected region.

\section{Ablation Studies}

We tested two aspects of DCSRL: the signals used to estimate exemplar
drift and the use of that estimate to weight replay losses. Four
variants each removed one drift signal. Two further variants disabled
drift conditioning for either the boundary or the structural loss
while retaining the underlying supervised loss. All variants used the
same task order, replay-memory policy, training settings, and random
seeds. Reliability-guided replay served as the control with both
conditioning branches disabled and is reported in
Table~\ref{tab:final_segmentation_all_methods}.

\paragraph{CAMUS.}
Full DCSRL achieved the highest mean Dice and boundary F1, although
Dice varied by only $0.0014$ across all variants
(Table~\ref{tab:dcsrl_ablation}). The surface-distance measures
favored different configurations. Removing probability drift yielded
the lowest HD95 ($8.1014$), while disabling boundary conditioning
yielded the lowest ASD ($2.8307$) and matched the full model's IoU
($0.8172$) at the reported precision.

\begin{table*}[t]
\centering
\caption{Component ablation results
of the DCSRL framework.}
\label{tab:dcsrl_ablation}
\resizebox{\textwidth}{!}{%
\begin{tabular}{llccccc}
\toprule
\textbf{Dataset} & \textbf{Variant} &
\textbf{Dice $\uparrow$} & \textbf{IoU $\uparrow$} &
\textbf{HD95 $\downarrow$} & \textbf{ASD $\downarrow$} &
\textbf{Boundary F1 $\uparrow$} \\
\midrule
\multirow{7}{*}{CAMUS}
& Full DCSRL
& $\mathbf{0.8986\pm0.0030}$ & $\mathbf{0.8172\pm0.0049}$
& $8.4169\pm0.7366$ & $2.8752\pm0.1583$ & $\mathbf{0.5900\pm0.0033}$ \\
& w/o Feature Drift
& $0.8972\pm0.0020$ & $0.8150\pm0.0032$
& $8.4092\pm0.3539$ & $2.8947\pm0.0699$ & $0.5864\pm0.0031$ \\
& w/o Probability Drift
& $0.8981\pm0.0021$ & $0.8165\pm0.0035$
& $\mathbf{8.1014\pm0.5967}$ & $2.8402\pm0.1590$ & $0.5861\pm0.0021$ \\
& w/o Boundary Drift
& $0.8980\pm0.0013$ & $0.8164\pm0.0022$
& $8.3509\pm0.1885$ & $2.8674\pm0.0482$ & $0.5881\pm0.0019$ \\
& w/o Structural Drift
& $0.8980\pm0.0019$ & $0.8164\pm0.0031$
& $8.1242\pm0.2546$ & $2.8364\pm0.0850$ & $0.5859\pm0.0065$ \\
& w/o Boundary Conditioning
& $0.8985\pm0.0020$ & $\mathbf{0.8172\pm0.0032}$
& $8.1326\pm0.5401$ & $\mathbf{2.8307\pm0.0869}$ & $0.5889\pm0.0063$ \\
& w/o Structural Conditioning
& $0.8982\pm0.0017$ & $0.8166\pm0.0028$
& $8.1778\pm0.2651$ & $2.8473\pm0.0720$ & $0.5872\pm0.0009$ \\
\midrule
\multirow{7}{*}{CardiacNet}
& Full DCSRL
& $\mathbf{0.8218\pm0.0029}$ & $\mathbf{0.6978\pm0.0041}$
& $\mathbf{10.2148\pm0.4996}$ & $\mathbf{3.7972\pm0.1541}$ & $0.5998\pm0.0071$ \\
& w/o Feature Drift
& $0.8217\pm0.0012$ & $0.6977\pm0.0015$
& $10.4531\pm0.9575$ & $3.8113\pm0.1986$ & $0.5983\pm0.0024$ \\
& w/o Probability Drift
& $0.8205\pm0.0009$ & $0.6960\pm0.0014$
& $10.5543\pm0.9134$ & $3.8761\pm0.2159$ & $0.5985\pm0.0038$ \\
& w/o Boundary Drift
& $0.8206\pm0.0020$ & $0.6961\pm0.0029$
& $10.6801\pm0.2106$ & $3.9065\pm0.0598$ & $0.5979\pm0.0042$ \\
& w/o Structural Drift
& $0.8211\pm0.0021$ & $0.6968\pm0.0031$
& $10.7225\pm0.8841$ & $3.9071\pm0.2008$ & $\mathbf{0.6000\pm0.0069}$ \\
& w/o Boundary Conditioning
& $0.8211\pm0.0050$ & $0.6969\pm0.0071$
& $10.2307\pm1.1434$ & $3.8003\pm0.2944$ & $0.5964\pm0.0079$ \\
& w/o Structural Conditioning
& $0.8213\pm0.0034$ & $0.6972\pm0.0046$
& $10.6956\pm1.3333$ & $3.8997\pm0.4195$ & $0.5992\pm0.0078$ \\
\midrule
\multirow{7}{*}{EchoNet-Dynamic}
& Full DCSRL
& $\mathbf{0.9148\pm0.0005}$ & $\mathbf{0.8429\pm0.0009}$
& $\mathbf{15.3023\pm0.2800}$ & $\mathbf{4.1372\pm0.0527}$
& $\mathbf{0.5283\pm0.0085}$ \\
& w/o Feature Drift
& $0.9110\pm0.0002$ & $0.8366\pm0.0004$
& $15.8196\pm0.4126$ & $4.3089\pm0.0726$ & $0.5150\pm0.0013$ \\
& w/o Probability Drift
& $0.9109\pm0.0004$ & $0.8363\pm0.0007$
& $15.7062\pm0.3316$ & $4.3013\pm0.0610$ & $0.5144\pm0.0004$ \\
& w/o Boundary Drift
& $0.9108\pm0.0007$ & $0.8363\pm0.0012$
& $15.8187\pm0.2762$ & $4.3191\pm0.0399$ & $0.5131\pm0.0023$ \\
& w/o Structural Drift
& $0.9110\pm0.0012$ & $0.8366\pm0.0021$
& $15.7303\pm0.5187$ & $4.3086\pm0.1202$ & $0.5140\pm0.0081$ \\
& w/o Boundary Conditioning
& $0.9109\pm0.0007$ & $0.8364\pm0.0012$
& $15.8913\pm0.3425$ & $4.3307\pm0.0521$ & $0.5135\pm0.0003$ \\
& w/o Structural Conditioning
& $0.9113\pm0.0009$ & $0.8370\pm0.0015$
& $15.5972\pm0.3332$ & $4.2753\pm0.0463$ & $0.5160\pm0.0008$ \\
\bottomrule
\end{tabular}%
}

\par\smallskip
\begin{minipage}{\linewidth}
\footnotesize
\textit{Note:} Results follow sequential
training on CAMUS, CardiacNet, and
EchoNet-Dynamic. Values are mean
$\pm$ sample standard deviation over
three seeds (42, 123, and 2025).
Higher Dice, IoU, and Boundary F1 are
better; lower HD95 and ASD are better.
Bold indicates the best mean within
each dataset, including ties at the
displayed precision.
\end{minipage}

\end{table*}

\paragraph{CardiacNet.}
Full DCSRL achieved the highest mean Dice and IoU and the lowest
HD95 and ASD. Disabling structural conditioning had a larger effect
on surface distance than on overlap: HD95 increased from $10.2148$
to $10.6956$ and ASD from $3.7972$ to $3.8997$, while Dice changed
from $0.8218$ to $0.8213$. Removing the structural drift signal
yielded the highest mean boundary F1 ($0.6000$), only $0.0002$
above the full model, and increased HD95 to $10.7225$.

\paragraph{EchoNet-Dynamic.}
The final task showed the most consistent response to ablation.
Removing any drift signal or disabling either conditioning branch
worsened all five mean metrics. Boundary F1 fell from $0.5283$ for
full DCSRL to $0.5131$--$0.5160$ across the six variants. Disabling
boundary conditioning produced the largest increases in both HD95
and ASD. Thus, the full configuration was favored across all measured
aspects of LV segmentation on EchoNet-Dynamic, while the preferred
configuration on CAMUS depended on the measure of contour agreement.

\section{Computational Efficiency and Memory Analysis}
\label{sec:computational_efficiency}

DCSRL and reliability-guided replay use the same segmentation network
with 7,762,597 trainable parameters. They also retain the same exemplars
under a common memory budget. DCSRL adds two training costs: storing
a source-task signature for each retained exemplar and comparing that
signature with the current model response during replay. At inference,
both methods use the same segmentation forward pass. The replay buffer
and reference signatures are needed only for continued training.

\begin{table*}[t]
\centering
\caption{Computational cost and replay-storage
requirements during sequential training.}
\label{tab:computational_efficiency}
\setlength{\tabcolsep}{4pt}
\renewcommand{\arraystretch}{1.08}
\resizebox{\textwidth}{!}{%
\begin{tabular}{llcccc}
\toprule
\textbf{Task} & \textbf{Method} & \textbf{Training time (s)} &
\textbf{Time overhead} & \textbf{Peak GPU memory (GB)} &
\textbf{Replay storage (MB)} \\
\midrule
\multirow{2}{*}{CAMUS}
& Reliability-guided replay & $773.91 \pm 11.35$ & -- &
$1.6624 \pm 0.0017$ & 1200.00 \\
& DCSRL & $795.70 \pm 2.69$ & $+2.82\%$ &
$1.6629 \pm 0.0010$ & 1301.60 \\
\midrule
\multirow{2}{*}{CardiacNet}
& Reliability-guided replay & $16286.35 \pm 74.41$ & -- &
$2.3086 \pm 0.0002$ & 1350.00 \\
& DCSRL & $16469.97 \pm 96.74$ & $+1.13\%$ &
$2.5320 \pm 0.0013$ & 1464.30 \\
\midrule
\multirow{2}{*}{EchoNet-Dynamic}
& Reliability-guided replay & $6715.91 \pm 74.12$ & -- &
$2.2707 \pm 0.0011$ & 1350.00 \\
& DCSRL & $7747.30 \pm 47.82$ & $+15.36\%$ &
$2.4943 \pm 0.0003$ & 1464.30 \\
\bottomrule
\end{tabular}%
}

\par\smallskip
\begin{minipage}{\linewidth}
\footnotesize
\textit{Note:} Training time and peak GPU
memory are mean $\pm$ sample standard
deviation over seeds 42, 123, and 2025.
Time overhead denotes the percentage
increase relative to reliability-guided
replay for each task.
Replay storage is the persistent footprint
after each task and includes source-task
reference signatures for DCSRL.
\end{minipage}

\end{table*}

\paragraph{Training time and peak GPU memory.}
Across the complete sequence, DCSRL increased mean training time from
6.60 to 6.95 hours, an additional 20.6 minutes ($5.20\%$).
EchoNet-Dynamic accounted for 17.2 minutes of this increase.
CardiacNet was the longest training stage for both methods, but its
relative time overhead was the smallest ($1.13\%$). Peak GPU memory
was nearly identical on CAMUS and increased by approximately
0.22~GB on each subsequent task. The highest mean peak for DCSRL
was 2.53~GB on CardiacNet as shown in Table~\ref{tab:computational_efficiency}.

\paragraph{Persistent replay storage.}
The buffer held 1,600 exemplars after CAMUS and reached its capacity
of 1800 after CardiacNet. At capacity, the source-task signatures
added 114.30~MB to the baseline footprint of 1350.00~MB, an increase
of $8.47\%$. The buffer remained at 1464.30~MB after EchoNet-Dynamic
as new exemplars replaced existing entries.

\section{Discussion}
\label{sec:discussion}

The comparison with reliability-guided replay shows that the supervision
applied to a stored exemplar can influence final segmentation quality
even when the network, memory capacity, and sampling policy are held
fixed. DCSRL increased the mean final Dice across datasets from $0.8718$
to $0.8784$. Its clearest benefit appeared on CardiacNet, where the model
was evaluated after subsequent training on EchoNet-Dynamic. Dice
increased from $0.8096$ to $0.8218$, while HD95 and ASD decreased by
$36.3\%$ and $27.1\%$, respectively
(Table~\ref{tab:final_segmentation_all_methods}). Thus, a modest change
in regional overlap accompanied a substantial reduction in the distance
between predicted and reference contours.

\paragraph{Final performance and transfer.}
The higher final Dice and the lower transfer scores describe different
aspects of the training trajectory. Backward transfer measures change
relative to each method's own performance immediately after learning a
task. DCSRL's near-zero backward transfer ($0.0003$) indicates that,
on average, its final scores on earlier tasks remained close to those
initial values. Reliability-guided replay showed a larger subsequent
gain ($0.0489$), even though its final Dice was lower on each dataset.
It also achieved higher forward transfer ($0.4366$ versus $0.4080$),
which in this study measures performance on a future dataset before
training on it (Table~\ref{tab:continual_learning_comparison}).
The evidence for DCSRL is therefore strongest for segmentation quality
at the end of the evaluated sequence.

\paragraph{What the ablations reveal.}
The contribution of drift conditioning depended on both the dataset and
the measure of contour agreement. On CardiacNet, disabling structural
conditioning worsened all five mean metrics. Removing the structural
drift signal produced a different outcome: boundary F1 increased by
only $0.0002$, while HD95 and ASD both increased
(Table~\ref{tab:dcsrl_ablation}). The distinction matters because
removing one input to the drift estimate still allows the remaining
signals to regulate structural supervision. On CAMUS, the full model
achieved the highest mean boundary F1, yet several reduced variants
had lower surface distances. Boundary F1 measures agreement within a
fixed tolerance, whereas HD95 captures the upper tail of contour
displacement. A gain in one can therefore coexist with a decline in
the other. With three training seeds, small differences between variants
warrant a descriptive interpretation.

EchoNet-Dynamic showed the most consistent ablation pattern: removing
any drift signal or disabling either conditioning branch worsened all
five mean metrics. Since EchoNet-Dynamic supplies LV-only supervision,
the structural replay term acts on eligible historical exemplars during
this stage. Its association with final LV performance suggests that the
way earlier examples are supervised can also influence learning on the
incoming domain.

\paragraph{Interpreting source-referenced drift.}
A frozen signature records what the model predicted when an exemplar
entered memory. A later departure from that prediction can reflect
either deterioration or correction of an earlier error. DCSRL uses
the departure to set the weight of the supervised replay loss, while
the stored annotation supplies the target. This distinction also
identifies a useful test of the mechanism: whether larger drift predicts
a subsequent increase in annotation-based error for the same exemplar.
Comparing DCSRL with a constant-weight control matched for average
supervision strength would further clarify how much of the gain comes
from assigning additional weight to particular exemplars.

\paragraph{Practical implications.}
The additional cost arises during model updating. Across the full
sequence, DCSRL increased training time by $5.20\%$ and required
$114.3$~MB of additional storage at the 1800-exemplar capacity
(Table~\ref{tab:computational_efficiency}). Both methods use the same
segmentation network at inference. The observed contour improvements
motivate a downstream evaluation of measurements such as ventricular
volume, for which the location of a segmentation error may matter as
much as its contribution to overall overlap.

\section{Limitations and Future Work}
\label{sec:limitations}

\paragraph{Task order and annotation scope.}
We evaluated three echocardiographic datasets in one order:
CAMUS $\rightarrow$ CardiacNet $\rightarrow$ EchoNet-Dynamic.
Because the LV-only task always appeared last, its position in the
sequence was coupled to a change in annotation scope. The present
design therefore cannot separate the effects of task order, image
distribution, and available labels. Repeating the experiment across
all six task permutations would test sensitivity to order. A separate
experiment using the same CAMUS images with either full or LV-only
annotations would isolate the effect of narrowing supervision while
holding image appearance fixed.

\paragraph{Interpretation of drift.}
A frozen source-task signature records the model's response when an
exemplar enters memory. A subsequent departure can reflect an emerging
segmentation error or a correction to an inaccurate source prediction.
The current drift score gives both changes a positive magnitude.
Comparing exemplar-level drift with changes in error against the
stored annotation would show when drift identifies deterioration that
warrants stronger supervision. A constant-weight control matched to
DCSRL's mean boundary and structural weights would also test how much
of the benefit comes from assigning supervision to particular
exemplars.

\paragraph{Sensitivity of the conditioning rule.}
The drift weights, normalization thresholds, and conditioning gains
were fixed across tasks. The ablations in
Table~\ref{tab:dcsrl_ablation} test the effect of removing individual
components, but leave open how sensitive performance is to their
weights. The structural descriptor also summarizes each class through
contour disagreement and connected-component count. A myocardial mask
can retain its connectivity while its wall thickness changes locally.
A useful extension would compare the current descriptor with a
wall-thickness or signed-distance measure on cases where the two
descriptors disagree.

\paragraph{Statistical and architectural scope.}
Three training seeds characterize run-to-run variation under the
chosen protocol. They provide limited evidence for ranking variants
with nearly identical means, and they do not quantify uncertainty
across patients or institutions. For example, removing structural
drift changed mean CardiacNet boundary F1 from $0.5998$ to $0.6000$.
Additional seeds and paired patient-level confidence intervals would
help distinguish training variability from variation across cases.
The experiments also used a single U-Net backbone. Since feature
drift is measured at its pooled bottleneck, testing a transformer
encoder under the same replay budget would show whether the current
normalization transfers to a different feature representation.

\paragraph{Replay storage and scale.}
DCSRL assumes continued access to stored images, annotations, and
source-task signatures. At the same 1800-exemplar capacity as
reliability-guided replay, the signatures increased persistent storage
by $8.47\%$. This leaves two practical questions. First, would DCSRL
retain its advantage under an equal storage budget in bytes, where
signature storage competes with space for additional exemplars?
Second, how would it perform over a longer sequence as a fixed buffer
must represent more domains? Evaluating both conditions on additional
institutions would test the method under tighter memory constraints
and a broader range of acquisition shifts.

\section{Conclusion}
\label{conclusion}

This study introduced Drift-Conditioned Structural Replay Loss (DCSRL) for continual echocardiographic segmentation under heterogeneous imaging distributions and annotation spaces. DCSRL builds upon a
reliability-guided replay strategy while introducing an exemplar-adaptive supervision mechanism that responds to changes in previously acquired knowledge. By comparing current model representations, class
probabilities, boundary responses, and predicted anatomical structures with frozen source-task references, DCSRL dynamically regulates boundary and structural preservation during sequential learning. Experimental evaluations across CAMUS, CardiacNet, and EchoNet-Dynamic demonstrated that drift-conditioned supervision improves final segmentation performance and anatomical boundary
preservation compared with conventional continual-learning approaches. The findings also revealed trade-offs between final segmentation accuracy and knowledge transfer, emphasizing that stronger anatomical preservation does not necessarily improve every aspect of continual
learning. Importantly, DCSRL achieves these improvements without
introducing additional trainable parameters or modifying the
inference-time architecture, although source-reference storage and
drift estimation increase training-time computational requirements. Future research will investigate the robustness of DCSRL across alternative task orders, imaging institutions, and annotation
configurations. Further work will also examine the relationship
between measured knowledge drift and subsequent segmentation errors
to determine when stronger anatomical supervision is beneficial.
Extending drift-conditioned replay to other medical imaging
modalities and investigating more efficient reference-storage
strategies could further improve its applicability to continual
medical image analysis.

\section*{Declaration of Generative AI and AI-assisted technologies in the writing process}
During the preparation of this work, the authors used ChatGPT to assist with language editing and improve the readability of the manuscript. After using this tool, the authors reviewed and edited the content as needed and take full responsibility for the content of the published article.

\section*{CRediT authorship contribution statement}
\textbf{Zahid Ullah:} Conceptualization, Methodology, Software, Formal analysis, Investigation, Data curation, Writing - original draft, Writing - review \& editing. \textbf{Jaeyoon Seo:} Writing - original draft, Writing - review \& editing. \textbf{Jihie Kim:} Formal analysis, Investigation, Supervision, Project administration, Project management.

\section*{\textbf{Declaration of Competing Interests}} The authors declare that they have no known competing financial interests or personal relationships that could have appeared to influence the work reported in this paper.

\section*{Data availability}

The datasets analyzed in this study were obtained from publicly available research resources. CAMUS is accessible at (\url{https://www.creatis.insa-lyon.fr/Challenge/camus/}). CardiacNet is available through (\url{https://github.com/xmed-lab/CardiacNet}), subject to the dataset provider's access instructions. EchoNet-Dynamic is available at (\url{https://echonet.github.io/dynamic/}), subject to registration and the applicable research-use agreement. No new clinical imaging dataset was collected for this study.


\section*{Acknowledgments}
{This work was supported by the Dongguk University Research Fund of 2026.
This research was supported by the MSIT(Ministry of Science and ICT), Korea, under the ITRC(Information Technology Research Center) support program(IITP-2026-RS-2020-II201789), and the Artificial Intelligence Convergence Innovation Human Resources Development(IITP-2026-RS-2023-00254592) supervised by the IITP(Institute for Information \& Communications Technology Planning \& Evaluation).}

\makeatletter
\renewcommand{\bibnumfmt}[1]{#1.\hspace{0.5em}}
\makeatother

\bibliography{sn-bibliography}

@article{leclerc2019deep,
  title={Deep learning for segmentation using an open large-scale dataset in 2D echocardiography},
  author={Leclerc, Sarah and Smistad, Erik and Pedrosa, Joao and {\O}stvik, Andreas and Cervenansky, Frederic and Espinosa, Florian and Espeland, Torvald and Berg, Erik Andreas Rye and Jodoin, Pierre-Marc and Grenier, Thomas and others},
  journal={IEEE transactions on medical imaging},
  volume={38},
  number={9},
  pages={2198--2210},
  year={2019},
  publisher={IEEE}
}

@inproceedings{yang2024cardiacnet,
  title={Cardiacnet: Learning to reconstruct abnormalities for cardiac disease assessment from echocardiogram videos},
  author={Yang, Jiewen and Lin, Yiqun and Pu, Bin and Guo, Jiarong and Xu, Xiaowei and Li, Xiaomeng},
  booktitle={European Conference on Computer Vision},
  pages={293--311},
  year={2024},
  organization={Springer}
}

@inproceedings{zenke2017continual,
  title={Continual learning through synaptic intelligence},
  author={Zenke, Friedemann and Poole, Ben and Ganguli, Surya},
  booktitle={International conference on machine learning},
  pages={3987--3995},
  year={2017},
  organization={Pmlr}
}

@article{chaudhry2019tiny,
  title={On tiny episodic memories in continual learning},
  author={Chaudhry, Arslan and Rohrbach, Marcus and Elhoseiny, Mohamed and Ajanthan, Thalaiyasingam and Dokania, Puneet K and Torr, Philip HS and Ranzato, Marc'Aurelio},
  journal={arXiv preprint arXiv:1902.10486},
  year={2019}
}

@article{ouyang2020video,
  title={Video-based AI for beat-to-beat assessment of cardiac function},
  author={Ouyang, David and He, Bryan and Ghorbani, Amirata and Yuan, Neal and Ebinger, Joseph and Langlotz, Curtis P and Heidenreich, Paul A and Harrington, Robert A and Liang, David H and Ashley, Euan A and others},
  journal={Nature},
  volume={580},
  number={7802},
  pages={252--256},
  year={2020},
  publisher={Nature Publishing Group UK, London}
}

@inproceedings{deng2024memsam,
  title={Memsam: Taming segment anything model for echocardiography video segmentation},
  author={Deng, Xiaolong and Wu, Huisi and Zeng, Runhao and Qin, Jing},
  booktitle={2024 IEEE/CVF Conference on Computer Vision and Pattern Recognition (CVPR)},
  pages={9622--9631},
  year={2024},
  organization={IEEE}
}

@article{wu2024continual,
  title={Continual learning in medical image analysis: A survey},
  author={Wu, Xinyao and Xu, Zhe and Tong, Raymond Kai-yu},
  journal={Computers in Biology and Medicine},
  volume={182},
  pages={109206},
  year={2024},
  publisher={Elsevier}
}

@article{de2021continual,
  title={A continual learning survey: Defying forgetting in classification tasks},
  author={De Lange, Matthias and Aljundi, Rahaf and Masana, Marc and Parisot, Sarah and Jia, Xu and Leonardis, Ale{\v{s}} and Slabaugh, Gregory and Tuytelaars, Tinne},
  journal={IEEE transactions on pattern analysis and machine intelligence},
  volume={44},
  number={7},
  pages={3366--3385},
  year={2021},
  publisher={IEEE}
}

@article{perkonigg2021dynamic,
  title={Dynamic memory to alleviate catastrophic forgetting in continual learning with medical imaging},
  author={Perkonigg, Matthias and Hofmanninger, Johannes and Herold, Christian J and Brink, James A and Pianykh, Oleg and Prosch, Helmut and Langs, Georg},
  journal={Nature communications},
  volume={12},
  number={1},
  pages={5678},
  year={2021},
  publisher={Nature Publishing Group UK, London}
}

@article{kirkpatrick2017overcoming,
  title={Overcoming catastrophic forgetting in neural networks},
  author={Kirkpatrick, James and Pascanu, Razvan and Rabinowitz, Neil and Veness, Joel and Desjardins, Guillaume and Rusu, Andrei A and Milan, Kieran and Quan, John and Ramalho, Tiago and Grabska-Barwinska, Agnieszka and others},
  journal={Proceedings of the national academy of sciences},
  volume={114},
  number={13},
  pages={3521--3526},
  year={2017},
  publisher={National Academy of Sciences}
}

@article{li2017learning,
  title={Learning without forgetting},
  author={Li, Zhizhong and Hoiem, Derek},
  journal={IEEE transactions on pattern analysis and machine intelligence},
  volume={40},
  number={12},
  pages={2935--2947},
  year={2017},
  publisher={IEEE}
}

@article{parisi2019continual,
  title={Continual lifelong learning with neural networks: A review},
  author={Parisi, German I and Kemker, Ronald and Part, Jose L and Kanan, Christopher and Wermter, Stefan},
  journal={Neural networks},
  volume={113},
  pages={54--71},
  year={2019},
  publisher={Elsevier}
}

@article{baweja2018towards,
  title={Towards Continual Learning in Medical Imaging},
  author={Baweja, Chaitanya and Glocker, Ben and Kamnitsas, Konstantinos},
  journal={arXiv preprint arXiv:1811.02496},
  year={2018}
}

@inproceedings{zhu2023continual,
  title={Continual semantic segmentation with automatic memory sample selection},
  author={Zhu, Lanyun and Chen, Tianrun and Yin, Jianxiong and See, Simon and Liu, Jun},
  booktitle={2023 IEEE/CVF Conference on Computer Vision and Pattern Recognition (CVPR)},
  pages={3082--3092},
  year={2023},
  organization={IEEE}
}

@inproceedings{michieli2021continual,
  title={Continual semantic segmentation via repulsion-attraction of sparse and disentangled latent representations},
  author={Michieli, Umberto and Zanuttigh, Pietro},
  booktitle={2021 IEEE/CVF Conference on Computer Vision and Pattern Recognition (CVPR)},
  pages={1114--1124},
  year={2021},
  organization={IEEE}
}

@article{lopez2017gradient,
  title={Gradient episodic memory for continual learning},
  author={Lopez-Paz, David and Ranzato, Marc'Aurelio},
  journal={Advances in neural information processing systems},
  volume={30},
  year={2017}
}

@article{ullah2026bbr,
  title={BBR-Net: Boundary-Balanced Replay for Continual Medical Image Segmentation},
  author={Ullah, Zahid and Choi, Sieun and Kim, Jihie},
  journal={arXiv preprint arXiv:2606.14731},
  year={2026}
}

@article{li2025clms,
  title={CLMS: Bridging domain gaps in medical imaging segmentation with source-free continual learning for robust knowledge transfer and adaptation},
  author={Li, Weilu and Zhang, Yun and Zhou, Hao and Yang, Wenhan and Xie, Zhi and He, Yao},
  journal={Medical Image Analysis},
  volume={100},
  pages={103404},
  year={2025},
  publisher={Elsevier}
}

@inproceedings{bera2023memory,
  title={Memory replay for continual medical image segmentation through atypical sample selection},
  author={Bera, Sutanu and Ummadi, Vinay and Sen, Debashis and Mandal, Subhamoy and Biswas, Prabir Kumar},
  booktitle={International conference on medical image computing and computer-assisted intervention},
  pages={513--522},
  year={2023},
  organization={Springer}
}

@inproceedings{ronneberger2015u,
  title={U-net: Convolutional networks for biomedical image segmentation},
  author={Ronneberger, Olaf and Fischer, Philipp and Brox, Thomas},
  booktitle={International Conference on Medical image computing and computer-assisted intervention},
  pages={234--241},
  year={2015},
  organization={Springer}
}

@article{oktay2018attention,
  title={Attention u-net: Learning where to look for the pancreas},
  author={Oktay, Ozan and Schlemper, Jo and Folgoc, Loic Le and Lee, Matthew and Heinrich, Mattias and Misawa, Kazunari and Mori, Kensaku and McDonagh, Steven and Hammerla, Nils Y and Kainz, Bernhard and others},
  journal={arXiv preprint arXiv:1804.03999},
  year={2018}
}

@article{chen2021transunet,
  title={Transunet: Transformers make strong encoders for medical image segmentation},
  author={Chen, Jieneng and Lu, Yongyi and Yu, Qihang and Luo, Xiangde and Adeli, Ehsan and Wang, Yan and Lu, Le and Yuille, Alan L and Zhou, Yuyin},
  journal={arXiv preprint arXiv:2102.04306},
  year={2021}
}

@inproceedings{rebuffi2017icarl,
  title={icarl: Incremental classifier and representation learning},
  author={Rebuffi, Sylvestre-Alvise and Kolesnikov, Alexander and Sperl, Georg and Lampert, Christoph H},
  booktitle={2017 IEEE conference on computer vision and pattern recognition (CVPR)},
  pages={5533--5542},
  year={2017},
  organization={IEEE}
}

@article{zhang2023s,
  title={S 3 R: Shape and semantics-based selective regularization for explainable continual segmentation across multiple sites},
  author={Zhang, Jingyang and Gu, Ran and Xue, Peng and Liu, Mianxin and Zheng, Hao and Zheng, Yefeng and Ma, Lei and Wang, Guotai and Gu, Lixu},
  journal={IEEE Transactions on Medical Imaging},
  volume={42},
  number={9},
  pages={2539--2551},
  year={2023},
  publisher={IEEE}
}

@inproceedings{ranem2024continual,
  title={Continual atlas-based segmentation of prostate MRI},
  author={Ranem, Amin and Gonz{\'a}lez, Camila and dos Santos, Daniel Pinto and Bucher, Andreas M and Othman, Ahmed E and Mukhopadhyay, Anirban},
  booktitle={2024 IEEE/CVF Winter Conference on Applications of Computer Vision (WACV)},
  pages={7548--7557},
  year={2024},
  organization={IEEE}
}

@article{kervadec2021boundary,
  title={Boundary loss for highly unbalanced segmentation},
  author={Kervadec, Hoel and Bouchtiba, Jihene and Desrosiers, Christian and Granger, Eric and Dolz, Jose and Ayed, Ismail Ben},
  journal={Medical image analysis},
  volume={67},
  pages={101851},
  year={2021},
  publisher={Elsevier}
}

@article{karimi2019reducing,
  title={Reducing the hausdorff distance in medical image segmentation with convolutional neural networks},
  author={Karimi, Davood and Salcudean, Septimiu E},
  journal={IEEE Transactions on medical imaging},
  volume={39},
  number={2},
  pages={499--513},
  year={2019},
  publisher={IEEE}
}

@article{lin2024boundary,
  title={Boundary-wise loss for medical image segmentation based on fuzzy rough sets},
  author={Lin, Qiao and Chen, Xin and Chen, Chao and Garibaldi, Jonathan M},
  journal={Information Sciences},
  volume={661},
  pages={120183},
  year={2024},
  publisher={Elsevier}
}

@article{rolnick2019experience,
  title={Experience replay for continual learning},
  author={Rolnick, David and Ahuja, Arun and Schwarz, Jonathan and Lillicrap, Timothy and Wayne, Gregory},
  journal={Advances in neural information processing systems},
  volume={32},
  year={2019}
}

@article{aljundi2019online,
  title={Online continual learning with maximal interfered retrieval},
  author={Aljundi, Rahaf and Belilovsky, Eugene and Tuytelaars, Tinne and Charlin, Laurent and Caccia, Massimo and Lin, Min and Page-Caccia, Lucas},
  journal={Advances in neural information processing systems},
  volume={32},
  year={2019}
}

@article{buzzega2020dark,
  title={Dark experience for general continual learning: a strong, simple baseline},
  author={Buzzega, Pietro and Boschini, Matteo and Porrello, Angelo and Abati, Davide and Calderara, Simone},
  journal={Advances in neural information processing systems},
  volume={33},
  pages={15920--15930},
  year={2020}
}

@article{caccia2021new,
  title={New insights on reducing abrupt representation change in online continual learning},
  author={Caccia, Lucas and Aljundi, Rahaf and Asadi, Nader and Tuytelaars, Tinne and Pineau, Joelle and Belilovsky, Eugene},
  journal={arXiv preprint arXiv:2104.05025},
  year={2021}
}

@article{kang2024advancing,
  title={Advancing medical image segmentation: morphology-driven learning with diffusion transformer},
  author={Kang, Sungmin and Song, Jaeha and Kim, Jihie},
  journal={arXiv preprint arXiv:2408.00347},
  year={2024}
}

@article{isensee2021nnu,
  title={nnU-Net: a self-configuring method for deep learning-based biomedical image segmentation},
  author={Isensee, Fabian and Jaeger, Paul F and Kohl, Simon AA and Petersen, Jens and Maier-Hein, Klaus H},
  journal={Nature methods},
  volume={18},
  number={2},
  pages={203--211},
  year={2021},
  publisher={Nature Publishing Group US, New York}
}

@inproceedings{cao2022swin,
  title={Swin-unet: Unet-like pure transformer for medical image segmentation},
  author={Cao, Hu and Wang, Yueyue and Chen, Joy and Jiang, Dongsheng and Zhang, Xiaopeng and Tian, Qi and Wang, Manning},
  booktitle={European conference on computer vision},
  pages={205--218},
  year={2022},
  organization={Springer}
}

@inproceedings{kirillov2023segment,
  title={Segment anything},
  author={Kirillov, Alexander and Mintun, Eric and Ravi, Nikhila and Mao, Hanzi and Rolland, Chloe and Gustafson, Laura and Xiao, Tete and Whitehead, Spencer and Berg, Alexander C and Lo, Wan-Yen and others},
  booktitle={2023 IEEE/CVF international conference on computer vision (ICCV)},
  pages={3992--4003},
  year={2023},
  organization={IEEE}
}

@article{ma2024segment,
  title   = {Segment anything in medical images},
  author  = {Ma, Jun and He, Yuting and Li, Feifei and Han, Lin and You, Chenyu and Wang, Bo},
  journal = {Nature Communications},
  volume  = {15},
  pages   = {654},
  year    = {2024}
}

@article{leclerc2020lu,
  title={LU-Net: a multistage attention network to improve the robustness of segmentation of left ventricular structures in 2-D echocardiography},
  author={Leclerc, Sarah and Smistad, Erik and {\O}stvik, Andreas and Cervenansky, Frederic and Espinosa, Florian and Espeland, Torvald and Berg, Erik Andreas Rye and Belhamissi, Mourad and Israilov, Sardor and Grenier, Thomas and others},
  journal={IEEE Transactions on Ultrasonics, Ferroelectrics, and Frequency Control},
  volume={67},
  number={12},
  pages={2519--2530},
  year={2020},
  publisher={IEEE}
}

@article{oktay2017anatomically,
  title={Anatomically constrained neural networks (ACNNs): application to cardiac image enhancement and segmentation},
  author={Oktay, Ozan and Ferrante, Enzo and Kamnitsas, Konstantinos and Heinrich, Mattias and Bai, Wenjia and Caballero, Jose and Cook, Stuart A and De Marvao, Antonio and Dawes, Timothy and O‘Regan, Declan P and others},
  journal={IEEE transactions on medical imaging},
  volume={37},
  number={2},
  pages={384--395},
  year={2017},
  publisher={IEEE}
}

@article{painchaud2020cardiac,
  title={Cardiac segmentation with strong anatomical guarantees},
  author={Painchaud, Nathan and Skandarani, Youssef and Judge, Thierry and Bernard, Olivier and Lalande, Alain and Jodoin, Pierre-Marc},
  journal={IEEE transactions on medical imaging},
  volume={39},
  number={11},
  pages={3703--3713},
  year={2020},
  publisher={IEEE}
}

@article{painchaud2022echocardiography,
  title={Echocardiography segmentation with enforced temporal consistency},
  author={Painchaud, Nathan and Duchateau, Nicolas and Bernard, Olivier and Jodoin, Pierre-Marc},
  journal={IEEE transactions on medical imaging},
  volume={41},
  number={10},
  pages={2867--2878},
  year={2022},
  publisher={IEEE}
}

@article{wang2024comprehensive,
  title={A comprehensive survey of continual learning: Theory, method and application},
  author={Wang, Liyuan and Zhang, Xingxing and Su, Hang and Zhu, Jun},
  journal={IEEE transactions on pattern analysis and machine intelligence},
  volume={46},
  number={8},
  pages={5362--5383},
  year={2024},
  publisher={IEEE}
}

@article{alzubaidi2021review,
  title={Review of deep learning: concepts, CNN architectures, challenges, applications, future directions},
  author={Alzubaidi, Laith and Zhang, Jinglan and Humaidi, Amjad J and Al-Dujaili, Ayad and Duan, Ye and Al-Shamma, Omran and Santamar{\'\i}a, Jos{\'e} and Fadhel, Mohammed A and Al-Amidie, Muthana and Farhan, Laith},
  journal={Journal of big Data},
  volume={8},
  number={1},
  pages={53},
  year={2021},
  publisher={Springer}
}

@article{gama2014survey,
  title={A survey on concept drift adaptation},
  author={Gama, Jo{\~a}o and {\v{Z}}liobait{\.e}, Indr{\.e} and Bifet, Albert and Pechenizkiy, Mykola and Bouchachia, Abdelhamid},
  journal={ACM computing surveys (CSUR)},
  volume={46},
  number={4},
  pages={1--37},
  year={2014},
  publisher={ACM, New York, NY, USA}
}

@article{lu2018learning,
  title={Learning under concept drift: A review},
  author={Lu, Jie and Liu, Anjin and Dong, Fan and Gu, Feng and Gama, Joao and Zhang, Guangquan},
  journal={IEEE transactions on knowledge and data engineering},
  volume={31},
  number={12},
  pages={2346--2363},
  year={2018},
  publisher={IEEE}
}

@inproceedings{aljundi2018memory,
  title={Memory aware synapses: Learning what (not) to forget},
  author={Aljundi, Rahaf and Babiloni, Francesca and Elhoseiny, Mohamed and Rohrbach, Marcus and Tuytelaars, Tinne},
  booktitle={European conference on computer vision},
  pages={144--161},
  year={2018},
  organization={Springer}
}

@inproceedings{michieli2019incremental,
  title={Incremental learning techniques for semantic segmentation},
  author={Michieli, Umberto and Zanuttigh, Pietro},
  booktitle={2019 IEEE/CVF International Conference on Computer Vision Workshop (ICCVW)},
  pages={3205--3212},
  year={2019},
  organization={IEEE}
}

@inproceedings{cermelli2020modeling,
  title={Modeling the background for incremental learning in semantic segmentation},
  author={Cermelli, Fabio and Mancini, Massimiliano and Bulo, Samuel Rota and Ricci, Elisa and Caputo, Barbara},
  booktitle={2020 IEEE/CVF Conference on Computer Vision and Pattern Recognition (CVPR)},
  pages={9230--9239},
  year={2020},
  organization={IEEE}
}

@inproceedings{douillard2021plop,
  title={Plop: Learning without forgetting for continual semantic segmentation},
  author={Douillard, Arthur and Chen, Yifu and Dapogny, Arnaud and Cord, Matthieu},
  booktitle={2021 IEEE/CVF Conference on Computer Vision and Pattern Recognition (CVPR)},
  pages={4039--4049},
  year={2021},
  organization={IEEE}
}

@inproceedings{ozdemir2018learn,
  title={Learn the new, keep the old: Extending pretrained models with new anatomy and images},
  author={Ozdemir, Firat and Fuernstahl, Philipp and Goksel, Orcun},
  booktitle={International Conference on Medical Image Computing and Computer-Assisted Intervention},
  pages={361--369},
  year={2018},
  organization={Springer}
}

@inproceedings{gonzalez2025wrong,
  title={What is wrong with continual learning in medical image segmentation?},
  author={Gonzalez, Camila and Lemke, Nick and Ranem, Amin and Sakas, Georgios and Mukhopadhyay, Anirban},
  booktitle={Proceedings of the International Workshop on Personalized Incremental Learning in Medicine},
  pages={25--34},
  year={2025}
}

@article{gonzalez2023lifelong,
  title={Lifelong nnU-Net: a framework for standardized medical continual learning},
  author={Gonz{\'a}lez, Camila and Ranem, Amin and Pinto dos Santos, Daniel and Othman, Ahmed and Mukhopadhyay, Anirban},
  journal={Scientific Reports},
  volume={13},
  number={1},
  pages={9381},
  year={2023},
  publisher={Nature Publishing Group UK, London}
}

@inproceedings{zhang2023continual,
  title={Continual learning for abdominal multi-organ and tumor segmentation},
  author={Zhang, Yixiao and Li, Xinyi and Chen, Huimiao and Yuille, Alan L and Liu, Yaoyao and Zhou, Zongwei},
  booktitle={International conference on medical image computing and computer-assisted intervention},
  pages={35--45},
  year={2023},
  organization={Springer}
}

@inproceedings{ji2023continual,
  title={Continual segment: Towards a single, unified and non-forgetting continual segmentation model of 143 whole-body organs in ct scans},
  author={Ji, Zhanghexuan and Guo, Dazhou and Wang, Puyang and Yan, Ke and Lu, Le and Xu, Minfeng and Wang, Qifeng and Ge, Jia and Gao, Mingchen and Ye, Xianghua and others},
  booktitle={2023 IEEE/CVF International Conference on Computer Vision (ICCV)},
  pages={21083--21094},
  year={2023},
  organization={IEEE}
}

@article{kiyasseh2021clinical,
  title={A clinical deep learning framework for continually learning from cardiac signals across diseases, time, modalities, and institutions},
  author={Kiyasseh, Dani and Zhu, Tingting and Clifton, David},
  journal={Nature Communications},
  volume={12},
  number={1},
  pages={4221},
  year={2021},
  publisher={Nature Publishing Group UK, London}
}

@inproceedings{lenga2020continual,
  title={Continual learning for domain adaptation in chest x-ray classification},
  author={Lenga, Matthias and Schulz, Heinrich and Saalbach, Axel},
  booktitle={Medical Imaging with Deep Learning},
  pages={413--423},
  year={2020},
  organization={PMLR}
}

@article{chaudhry2018efficient,
  title={Efficient lifelong learning with a-gem},
  author={Chaudhry, Arslan and Ranzato, Marc'Aurelio and Rohrbach, Marcus and Elhoseiny, Mohamed},
  journal={arXiv preprint arXiv:1812.00420},
  year={2018}
}

@article{shin2017continual,
  title={Continual learning with deep generative replay},
  author={Shin, Hanul and Lee, Jung Kwon and Kim, Jaehong and Kim, Jiwon},
  journal={Advances in neural information processing systems},
  volume={30},
  year={2017}
}

@article{aljundi2019gradient,
  title={Gradient based sample selection for online continual learning},
  author={Aljundi, Rahaf and Lin, Min and Goujaud, Baptiste and Bengio, Yoshua},
  journal={Advances in neural information processing systems},
  volume={32},
  year={2019}
}

@inproceedings{bang2021rainbow,
  title={Rainbow memory: Continual learning with a memory of diverse samples},
  author={Bang, Jihwan and Kim, Heesu and Yoo, YoungJoon and Ha, Jung-Woo and Choi, Jonghyun},
  booktitle={2021 IEEE/CVF conference on computer vision and pattern recognition (CVPR)},
  pages={8214--8223},
  year={2021},
  organization={IEEE}
}

@inproceedings{prabhu2020gdumb,
  title={Gdumb: A simple approach that questions our progress in continual learning},
  author={Prabhu, Ameya and Torr, Philip HS and Dokania, Puneet K},
  booktitle={European conference on computer vision},
  pages={524--540},
  year={2020},
  organization={Springer}
}

@article{boschini2022class,
  title={Class-incremental continual learning into the extended der-verse},
  author={Boschini, Matteo and Bonicelli, Lorenzo and Buzzega, Pietro and Porrello, Angelo and Calderara, Simone},
  journal={IEEE transactions on pattern analysis and machine intelligence},
  volume={45},
  number={5},
  pages={5497--5512},
  year={2022},
  publisher={IEEE}
}

@article{arani2022learning,
  title={Learning fast, learning slow: A general continual learning method based on complementary learning system},
  author={Arani, Elahe and Sarfraz, Fahad and Zonooz, Bahram},
  journal={arXiv preprint arXiv:2201.12604},
  year={2022}
}

@article{ma2021loss,
  title={Loss odyssey in medical image segmentation},
  author={Ma, Jun and Chen, Jianan and Ng, Matthew and Huang, Rui and Li, Yu and Li, Chen and Yang, Xiaoping and Martel, Anne L},
  journal={Medical image analysis},
  volume={71},
  pages={102035},
  year={2021},
  publisher={Elsevier}
}

@article{hu2019topology,
  title={Topology-preserving deep image segmentation},
  author={Hu, Xiaoling and Li, Fuxin and Samaras, Dimitris and Chen, Chao},
  journal={Advances in neural information processing systems},
  volume={32},
  year={2019}
}

@article{clough2020topological,
  title={A topological loss function for deep-learning based image segmentation using persistent homology},
  author={Clough, James R and Byrne, Nicholas and Oksuz, Ilkay and Zimmer, Veronika A and Schnabel, Julia A and King, Andrew P},
  journal={IEEE transactions on pattern analysis and machine intelligence},
  volume={44},
  number={12},
  pages={8766--8778},
  year={2020},
  publisher={IEEE}
}

@inproceedings{shit2021cldice,
  title={clDice-a novel topology-preserving loss function for tubular structure segmentation},
  author={Shit, Suprosanna and Paetzold, Johannes C and Sekuboyina, Anjany and Ezhov, Ivan and Unger, Alexander and Zhylka, Andrey and Pluim, Josien PW and Bauer, Ulrich and Menze, Bjoern H},
  booktitle={2021 IEEE/CVF Conference on Computer Vision and Pattern Recognition (CVPR)},
  pages={16555--16564},
  year={2021},
  organization={IEEE}
}

\end{document}